\documentclass{article}
\usepackage{blindtext}
\usepackage[T1]{fontenc}
\usepackage[utf8]{inputenc}

\usepackage{amssymb,chicago}
\usepackage{amsfonts}
\usepackage{amsmath}
\usepackage{hyperref,color}
\usepackage{amsthm}
\usepackage{geometry}
\usepackage{indentfirst}
\usepackage{bbm}

\newcommand{\eref}[1]{(\ref{#1})}
\usepackage[]{graphicx}
\usepackage{
tikz,
relsize,
booktabs,
tikz
}

\pgfdeclarelayer{background layer}
\pgfdeclarelayer{foreground layer}
\pgfsetlayers{background layer,main,foreground layer}

\tikzset{
>=latex
}
\usepackage{bbm}
\usepackage{appendix}
\usepackage{etoolbox}

\usepackage{etoc}

\newtheorem{mydef}{Preliminary}

\newtheorem{prop}{Proposition}
\newtheorem{rem}{Remark}

\newcommand{\cH}{\mathcal{H}} 
\newcommand{\cP}{\mathcal{P}} 
\newcommand{\cF}{\mathcal{F}} 

\newcommand{\pkg}[1]{{\normalfont\fontseries{b}\selectfont #1}}

\let\fct=\texttt

\DeclareMathOperator*{\argmin}{arg\,min}

\newenvironment{example}
{\begin{exemp}\begin{em}}
{\end{em}\end{exemp}}
\newtheorem{exemp}{Example}

\usepackage{algorithm}
\usepackage{algpseudocode}
\usepackage{aecompl}
\usepackage{array}
\usepackage{multirow}

\begin{document} 

\title{\bf \pkg{RKHSMetaMod}: An R package to estimate the Hoeffding decomposition of a complex model by solving RKHS ridge group sparse optimization problem\thanks{Halaleh Kamari, Universit\'e Paris-Saclay, France, {\sf @}, Sylvie Huet, INRAE, France, {\sf @}, Marie-Luce Taupin, UniversitÃ© Evry Val d'Essonne, France, {\sf @}.}}

\author{{\sc Halaleh Kamari},
{\sc Sylvie Huet},
{\sc Marie-Luce Taupin}
}
\maketitle

\begin{abstract} 
In this paper, we propose an R package, called \pkg{RKHSMetaMod}, that implements a procedure for estimating a meta-model of a complex model. The meta-model approximates the Hoeffding decomposition of the complex model and allows us to perform sensitivity analysis on it. It belongs to a reproducing kernel Hilbert space that is constructed as a direct sum of Hilbert spaces.
%In this paper, we propose an R package, called \pkg{RKHSMetaMod}, that implements a procedure allowing to approximate a complex model by an estimated meta-model. The meta-model belongs to a reproducing kernel Hilbert space which is constructed as a direct sum of Hilbert spaces. It approximates the Hoeffding decomposition of the complex model and allows us to perform sensitivity analysis on it.
The estimator of the meta-model is the solution of a penalized empirical least-squares minimization with the sum of the Hilbert norm and the empirical $L^2$-norm. This procedure, called RKHS ridge group sparse, allows both to select and estimate the terms in the Hoeffding decomposition, and therefore, to select and estimate the Sobol indices that are non-zero. The \pkg{RKHSMetaMod} package provides an interface from R statistical computing environment to the C++ libraries \pkg{Eigen} and \pkg{GSL}. In order to speed up the execution time and optimize the storage memory, except for a function that is written in R, all of the functions of this package are written using the efficient C++ libraries through \pkg{RcppEigen} and \pkg{RcppGSL} packages. These functions are then interfaced in the R environment in order to propose a user-friendly package.
\end{abstract}

\noindent
{\em Keywords:}
{Meta model},
{Hoeffding decomposition},
{Ridge Group Sparse penalty},
{Reproducing Kernel Hilbert Spaces}.

\section{Introduction}
Consider a phenomenon described by a model $m$ depending on $d$ input variables $X=(X_1,...,X_d)$. This model $m$ from $\mathbb{R}^d$ to $\mathbb{R}$ may be a known model that is calculable in all points of $X$, i.e. $Y=m(X)$, or it may be an unknown regression model defined as follows:
\begin{align}
\label{grm}
Y=m(X)+\varepsilon,
\end{align}
where the error $\varepsilon$ is assumed to be centered with a finite variance, i.e. $E(\varepsilon)=0$ and $\mbox{var}(\varepsilon)<\infty$. %, and independent of $X$. 
The components of $X$ are independent with a known law $P_X=\prod_{a=1}^dP_{X_a}$ on $\mathcal{X}$, a subset of $\mathbb{R}^d$. The number $d$ of components of $X$ may be large. The model $m$ may present high complexity as strong non-linearities and high order interaction effects, and it is assumed to be square-integrable, i.e. $m\in L^2(\mathcal{X},P_X)$.   
%Let $\{(X_i,Y_i)\}_{i=1}^n$ be $n$ data points. In this paper, we propose an R package that allows us to estimate a meta-model that approximates the Hoeffding decomposition of $m$ based on the data points.
Based on the data points $\{(X_i,Y_i)\}_{i=1}^n$, we estimate a meta-model that approximates the Hoeffding decomposition of $m$. This meta-model belongs to a reproducing kernel Hilbert space (RKHS), which is constructed as a direct sum of the Hilbert spaces leading to an additive decomposition including variables and interactions between them \cite{DURRANDE201357}. 
The estimator of the meta-model is calculated by minimizing an empirical least-squares criterion penalized by the sum of two penalty terms: the Hilbert norm and the empirical norm \cite{huet:hal-01434895}. This procedure allows us to select the subsets of variables $X_1,...,X_d$ that contribute to predict $Y$. The estimated meta-model is used to perform sensitivity analysis, and therefore, to determine the influence of each variable and groups of them on the output variable $Y$.  

%In the classical framework of sensitivity analysis, $m(X)$ is calculable in all points of $X$. In this framework, one may use the method of \citet{Sobol1993SensitivityEF} for variance-based methods of global sensitivity analysis in order to perform sensitivity analysis on $m$. 
In the classical framework of sensitivity analysis, the function $m$ is calculable in all points of $X$, and one may use the method of \cite{Sobol1993SensitivityEF} to perform sensitivity analysis on $m$. Let us briefly introduce this method:  
The independency between the components of $X$ leads to write the function $m$ according to its Hoeffding decomposition (\cite{Sobol1993SensitivityEF}, \cite{vaart_1998}):
\begin{align}
\label{sobol1}
m(X)=m_0+\sum_{a=1}^dm_a(X_a)+\sum_{a<a'}m_{a,a'}(X_a,X_{a'})+...+m_{1,...,d}(X).
\end{align}
The terms in this decomposition are defined using the conditional expected values:
\begin{align*}
m_0&=E_X(m(X)),\;
m_a(X_a)=E_X(m(X)|X_a)-m_0; \\
m_{a,a'}(X_a,X_{a'})&=E_X(m(X)|X_a,X_{a'})-m_a(X_a)-m_{a'}(X_{a'})-m_0, \cdots
\end{align*}
These terms are known as the constant term, main effects, interactions of order two and so on.
Let $\mathcal{P}$ be the set of all subsets of $\{1,...,d\}$ with dimension $1$ to $d$.  
For all $v\in\mathcal{P}$ and $X\in\mathcal{X}$, let $X_v$ be the vector with components $X_a$, $a\in v$. For a set $A$ let $\vert A\vert$ be its cardinality, and for all $v\in\mathcal{P}$, let $m_v:\mathbb{R}^{\vert v\vert}\rightarrow \mathbb{R}$ be the function associated with $X_v$ in Equation \eref{sobol1}. Then Equation \eref{sobol1} can be expressed as follows:
\begin{align}
\label{ch3sobol}
m(X)=m_0+\sum_{v\in\mathcal{P}}m_v(X_v).
\end{align} 
This decomposition is unique, all the terms $m_v$, $v\in\mathcal{P}$ are centered, and they are orthogonal with respect to $L^2(\mathcal{X},P_X)$.
The functions $m$ and $m_v$, $v\in\mathcal{P}$ in Equation \eref{ch3sobol} are square-integrable. As any two terms of decomposition  \eref{ch3sobol} are orthogonal, by squaring \eref{ch3sobol} and integrating it with respect to the distribution of $X$, a decomposition of the variance of $m(X)$ is obtained as follows:
\begin{align}
\label{chpacvardecop}
\text{var}(m(X))=\sum_{v\in\mathcal{P}}\text{var}(m_v(X_v)).
\end{align}
The Sobol indices associated with the group of variables $X_v$, $v \in \mathcal{P}$ are defined by:
\begin{align}
\label{trusob}
S_v=\frac{\text{var}(m_v(X_v))}{\text{var}(m(X))}.
\end{align}
For each $v$, $S_v$ expresses the fraction of variance of $m(X)$ explained by $X_v$. For all $v\in\mathcal{P}$, when $\vert v\vert=1$, the $S_v$s are referred to as the first order indices; when $\vert v\vert =2$, i.e. $v=\{a,a'\}$ and $a\neq a'$, they are referred to as the second order indices or the interaction indices of order two (between $X_a$ and $X_{a'}$); and the same holds for $\vert v\vert>2$.

The total number of the Sobol indices to be calculated is equal to $\vert\mathcal{P}\vert=2^d-1$, which raises exponentially with the number $d$ of the input variables. When $d$ is large, the evaluation of all the indices can be computationally demanding and even not reachable. 
%For this reason, only the indices of order not higher than two are calculated in practice.
In practice, only the indices of order not higher than two are calculated.  
However, only the first and second order indices may not provide a good information on the model sensitivities. In order to provide better information on the model sensitivities, \cite{HOMMA19961} proposed to calculate the first order and the total indices defined as follows:\\
Let $\mathcal{P}_a\subset\mathcal{P}$ be the set of all the subsets of $\{1,...,d\}$ including $a$, then $S_{T_a}=\sum_{v\in\mathcal{P}_a}S_v$.
For all $a\in\{1,...,d\}$, $S_{T_a}$ denotes the total effect of $X_a$. It expresses the fraction of variance of $m(X)$ explained by $X_a$ alone and all the interactions of it with the other variables.
The total indices allow us to rank the input variables with respect to the amount of their effect on the output variable. However, they do not provide complete information on the model sensitivities as do all the Sobol indices.

The classical computation of the Sobol indices is based on the Monte Carlo methods (see for example: \cite{Sobol1993SensitivityEF} for the main effect and interaction indices, and \cite{SALTELLI2002280} for the main effect and total indices). For models that are expensive to evaluate, the Monte Carlo methods lead to a high computational burden. Moreover, in the case where $d$ is large, $m$ is complex and the calculation of the variances (see Equation \eref{chpacvardecop}) is numerically complicated or not possible (as in the case where the model $m$ is unknown) the methods described above are not applicable. 
%Another method is to approximate $m$ by a simplified model, called a meta-model, which is much faster to evaluate and to perform sensitivity analysis on it. A meta-model provides additional information than just scalar indices.  
%Another method is to approximate $m$ by a simplified model, called a meta-model, and to perform sensitivity analysis on it. A meta-model which is much faster to evaluate, provides additional information than just scalar indices. It provides the approximations of the Sobol indices of $m$ at a lower computational cost and also a deeper view of the input variables effects on the model output. 
Another approach consists in approximating $m$ by a simplified model, called a meta-model, which is much faster to evaluate and to perform sensitivity analysis on it. Beside the approximations of the Sobol indices of $m$ at a lower computational cost, a meta-model provides a deeper view of the input variables effects on the model output.  
Among the meta-modelling methods proposed in the literature, the expansion based on polynomial Chaos (\cite{10.2307/2371268}, \cite{schoutens2000stochastic}) can be used to approximate the Hoeffding decomposition of $m$ \cite{SUDRET2008964}. 
The principle of the polynomial Chaos is to project $m$ onto a basis of orthonormal polynomials. The polynomial Chaos expansion of $m$ is written as \cite{doi:10.1137/S1064827503424505}:
\begin{align}
\label{ch3pc}
m(X)=\sum_{j=0}^\infty h_j\phi_j(X),
\end{align} 
where $\{h_j\}_{j=0}^\infty$ are the coefficients, and $\{\phi_j\}_{j=0}^\infty$ are the multivariate orthonormal polynomials associated with $X$ which are determined according to the distribution of the components of $X$. In practice, expansion \eref{ch3pc} shall be truncated for computational purposes, and the model $m$ may be approximated by $\sum_{j=0}^{v_{max}} h_j\phi_j(X)$, where $v_{max}$ is determined using a \textit{truncation scheme}. %In this approach, the Sobol indices are obtained by summing up the squares of the suitable coefficients.
The Sobol indices are obtained then by summing up the squares of the suitable coefficients. \cite{BlatmanSudret} proposed a method for truncating the polynomial Chaos expansion and an algorithm based on the least angle regression for selecting the terms in the expansion.
In this approach, according to the distribution of the components of $X$, a unique family of orthonormal polynomials $\{\phi_j\}_{j=0}^\infty$ is determined. However, this family may not be necessarily the best functional basis to approximate $m$ well. %Finally, the Sobol indices are obtained by summing up the squares of the suitable coefficients.  
%\citet{BlatmanSudret} proposed a method for truncating the polynomial Chaos expansion and an algorithm based on the least angle regression for selecting the terms in the expansion.
%In this method, according to the distribution of the components of $X$, a unique family of orthonormal polynomials $\{\phi_j\}_{j=0}^\infty$ is determined. However, this family may not be necessarily the best functional basis to approximate $m$ well.  
 
%Another method to construct meta-models is the Gaussian Process (GP) modelling (\citet{10.2307/1269548}, \citet{doi:10.1111/j.1467-9868.2004.05304.x}, \citet{Kleijnen:2007:DAS:1554802,KLEIJNEN2009707}, \citet{MARREL2009742}, \citet{Durrande2012}, \citet{doi:10.1137/130926869}). 
Gaussian Process (GP) can also be used to construct meta-models as highlighted in \cite{10.2307/1269548}, \cite{doi:10.1111/j.1467-9868.2004.05304.x}, \cite{Kleijnen:2007:DAS:1554802,KLEIJNEN2009707}, \cite{MARREL2009742}, \cite{Durrande2012}, \cite{doi:10.1137/130926869}.
The principle is to consider that the prior knowledge about the function $m(X)$, can be modelled by a GP $\mathcal{Z}(X)$ with a mean $m_{\mathcal{Z}}(X)$ and a covariance kernel $k_{\mathcal{Z}}(X,X')$.   
To perform sensitivity analysis from a GP model, one may replace the true model $m(X)$ with the mean of the conditional GP and deduce the Sobol indices from it. 
A review on the meta-modelling based on polynomial Chaos and GP is presented in \cite{LeGratiet2017}. 

\cite{DURRANDE201357} considered a class of the functional approximation methods similar to the GP and obtained a meta-model that satisfies the properties of the Hoeffding decomposition. %They proposed to approximate $m$ by functions belonging to a RKHS $\mathcal{H}$ which is a direct sum of the Hilbert spaces and is constructed in a way that the projection of $m$ onto $\mathcal{H}$, denoted $f^*$, is an approximation of the Hoeffding decomposition of $m$.  
They proposed to approximate $m$ by functions belonging to a RKHS $\mathcal{H}$ which is a direct sum of the Hilbert spaces. Their RKHS $\mathcal{H}$ is constructed in a way that the projection of $m$ onto $\mathcal{H}$, denoted $f^*$, is an approximation of the Hoeffding decomposition of $m$.  
The function $f^*$ is defined as the minimizer over the functions $f\in\mathcal{H}$ of the criterion $E_X(m(X)-f(X))^2.$

Let $<.,.>_{\mathcal{H}}$ be the scalar product in $\mathcal{H}$, let also $k$ and $k_v$ be the reproducing kernels associated with the RKHS $\mathcal{H}$ and the RKHS $\mathcal{H}_v$, respectively. The properties of the RKHS $\mathcal{H}$ insures that any function $f\in\mathcal{H}$, $f:\mathcal{X}\subset \mathbb{R}^d\rightarrow\mathbb{R}$ is written as the following decomposition: 
\begin{align}
\label{durandhoeff}
f(X)=<f,k(X,.)>_{\mathcal{H}}=f_0+\sum_{v\in\mathcal{P}}f_v(X_v),
\end{align}
where $f_0$ is constant, and $f_v:\mathbb{R}^{\vert v\vert}\rightarrow \mathbb{R}$ is defined by $f_v(X)=<f,k_v(X,.)>_{\mathcal{H}}$. For more details on the RKHS construction and the definition of the Hilbert norm see Section \ref{subsec:Data} in \ref{app:technical}. 

For all $v\in\mathcal{P}$, the functions $f_v(X_v)$ are centered and for all $v\neq v'$, the functions $f_v(X_v)$ and $f_{v'}(X_{v'})$ are orthogonal with respect to $L^2(\mathcal{X},P_X)$. Therefore, the decomposition of the function $f$ presented in Equation (\ref{durandhoeff}) is its Hoeffding decomposition. 
%As the function $f^*$ belongs to the RKHS $\mathcal{H}$, it is decomposed as its Hoeffding decomposition:
As the function $f^*$ belongs to the RKHS $\mathcal{H}$, it is decomposed as its Hoeffding decomposition, $f^*=f^*_0+\sum_{v\in\mathcal{P}}f^*_v$, and each function $f^*_v$ approximates the function $m_v$ in Equation \eref{ch3sobol}.
%\begin{equation}
%\label{ch3decfet}
%f^*=f^*_0+\sum_{v\in\mathcal{P}}f^*_v,
%\end{equation}
%and each function $f^*_v$ approximates the function $m_v$ in Equation \eref{ch3sobol}.  
%As mentioned earlier, in decomposition \eref{ch3decfet}, we have $\vert\mathcal{P}\vert=2^d-1$ terms $f^*_v$ to be estimated, which may be huge since it rises very quickly by increasing $d$.
The number of terms $f^*_v$ that should be estimated in the Hoeffding decomposition of $f^*$ is equal to $\vert\mathcal{P}\vert=2^d-1$, which may be huge since it rises very quickly by increasing $d$. 
In order to deal with this problem, in the regression framework, one may estimate $f^*$ by a sparse meta-model $\widehat{f}\in\mathcal{H}$. To this end, the estimation of $f^*$ is done on the basis of $n$ observations by minimizing a least-squares criterion suitably penalized in order to deal with both the non-parametric nature of the problem and the possibly large number of functions that have to be estimated. 
%Note that in the classical framework of the sensitivity analysis, where $m(X)$ is calculable in all points $X$, one may calculate a sparse approximation of $f^*$ using least-squares penalized criterion as it is done in the non-parametric regression framework. 
In the classical framework of the sensitivity analysis one may calculate a sparse approximation of $f^*$ using least-squares penalized criterion as it is done in the non-parametric regression framework. 
In order to obtain a sparse solution of a minimization problem, the penalty function should enforce the sparsity. 
There exists various ways of enforcing sparsity for a minimization (maximization) problem, see for example \cite{Hastie:2015:SLS:2834535} for a review. Some methods, such as the Sparse Additive Models (SpAM) procedure (\cite{Ravikumar}, \cite{NIPS2008_3616}) are based on a combination of the $l_1$-norm with the empirical $L^2$-norm: 
%\begin{align*}
%\Vert f\Vert_{n,1}=\sum_{a=1}^d\Vert f_a\Vert_n,
%\end{align*}
%where $$\Vert f_a\Vert_n^2=\frac{1}{n}\sum_{i=1}^nf_a^2(X_{ai}),$$ is the squared empirical $L^2$-norm of the univariate function $f_a$.
$\Vert f\Vert_{n,1}=\sum_{a=1}^d\Vert f_a\Vert_n,$ where $\Vert f_a\Vert_n^2=\sum_{i=1}^nf_a^2(X_{ai})/n,$ is the squared empirical $L^2$-norm of the univariate function $f_a$.
The Component Selection and Smoothing Operator (COSSO) method developed by \cite{lin2006} enforces sparsity using a combination of the $l_1$-norm with the Hilbert norm: $\Vert f\Vert_{\mathcal{H},1}=\sum_{a=1}^d\Vert f_a\Vert_{\mathcal{H}_a}$.
Instead of focusing on only one penalty term, one may consider a more general family of estimators, called \textit{doubly penalized estimator}, which is obtained by minimizing a criterion penalized by the sum of two penalty terms. \cite{NIPS2009_3688,Raskutti:2012:MRS:2503308.2188398} proposed a \textit{doubly penalized estimator}, which is the solution of the minimization of a least-squares criterion penalized by the sum of a sparsity penalty term and a combination of the $l_1$-norm with the Hilbert norm:
\begin{align}
\label{doubpen}
\gamma\Vert f\Vert_{n,1}+\mu\Vert f\Vert_{\mathcal{H},1},
\end{align}
where $\gamma, \mu\in\mathbb{R}$ are the tuning parameters that should be suitably chosen.
 
\cite{meier2009} proposed a related family of estimators, based on the penalization with the empirical $L^2$-norm. Their penalty function is the sum of the sparsity penalty term, $\Vert f\Vert_{n,1}$, and a smoothness penalty term. 
\cite{huet:hal-01434895} considered the same approximation functional spaces as \cite{DURRANDE201357}, and obtained a \textit{doubly penalized estimator} of a meta-model which approximates the Hoeffding decomposition of $m$. Their estimator is the solution of the least-squares minimization penalized by the penalty function defined in Equation \eref{doubpen} adapted to the multivariate setting,
\begin{align}
\label{ch3pen}
\gamma\Vert f\Vert_{n}+\mu\Vert f\Vert_{\mathcal{H}},\mbox{ with }\Vert f\Vert_{n}=\sum_{v\in\mathcal{P}}\Vert f_v\Vert_{n},\:\Vert f\Vert_{\mathcal{H}}=\sum_{v\in\mathcal{P}}\Vert f_v\Vert_{\mathcal{H}_v}.
\end{align}
%with $\Vert f\Vert_{n}=\sum_{v\in\mathcal{P}}\Vert f_v\Vert_{n}$ and $\Vert f\Vert_{\mathcal{H}}=\sum_{v\in\mathcal{P}}\Vert f_v\Vert_{\mathcal{H}_v}.$
%\begin{align*}
%\Vert f\Vert_{n}=\sum_{v\in\mathcal{P}}\Vert f_v\Vert_{n}
%\quad\text{and}\quad 
%\Vert f\Vert_{\mathcal{H}}=\sum_{v\in\mathcal{P}}\Vert f_v\Vert_{\mathcal{H}_v}.
%\end{align*}
This procedure, called RKHS ridge group sparse, estimates the groups $v$ that are suitable for predicting $f^*$, and the relationship between $f^*_v$ and $X_v$ for each group. 
The obtained estimator, called RKHS meta-model, is used then to estimate the Sobol indices of $m$. This approach renders it possible to estimate the Sobol indices for all groups in the support of the RKHS meta-model, including the interactions of possibly high order, a point known to be difficult in practice. 

In this paper, we introduce an R package, called \pkg{RKHSMetaMod}, that implements the RKHS ridge group sparse procedure. %This package deals with the input variables $X=(X_1,...,X_d)$ that are independent and uniformly distributed on $\mathcal{X}=[0,1]^d$, i.e. $X\sim P_X=P_1\times ...\times P_d$, with $P_a$, $a=1,...,d$ being the uniform law on the interval $[0,1]$. It allows us to: 
The functions of this package allows us to:
\begin{itemize}
\item[(1)] calculate the reproducing kernels and their associated Gram matrices (see Section \ref{subsec:Gramm}), 
\item[(2)] implement the RKHS ridge group sparse procedure and a special case of it, called the RKHS group lasso procedure (when $\gamma=0$ in the penalty function \eref{ch3pen}), in order to estimate the terms $f^*_v$ in the Hoeffding decomposition of the meta-model $f^*$ leading to an estimation of the function $m$ (see Section \ref{subsec:optim}), 
\item[(3)] choose the tuning parameters $\mu$ and $\gamma$ (see Equation \eref{ch3pen}), using a procedure that leads to obtain the \textit{best} RKHS meta-model in terms of the prediction quality,
\item[(4)] estimate the Sobol indices of the function $m$ (see Section \ref{subsec:SI}).
\end{itemize}
The current version of the package supports uniformly distributed input variables on $\mathcal{X}=[0,1]^d$. However, it could easily be adapted to a different distribution of input variables by making a small modification to one of its functions (see Remark \ref{ch3:remxun} of Section \ref{subsec:Gramm}).

Let us give a brief overview of the related existing statistical packages to \pkg{RKHSMetaMod}.
%Related existing statistical packages to \pkg{RKHSMetaMod} include:
%\begin{itemize}
%\item the packages which implement the sensitivity analysis methods such as \pkg{SAFE} (\citet{PIANOSI201580}) and \pkg{sensitivity} (\citet{sensitivity}). 
%The packages such as \pkg{SAFE} (\citet{PIANOSI201580}) and \pkg{sensitivity} (\citet{sensitivity}) are devoted to the implementation of the sensitivity analysis methods. 
%\pkg{SAFE} (Sensitivity Analysis For Everybody) is a Matlab toolbox which concerns the application of global sensitivity analysis and implements several methods of global sensitivity analysis including the method of \citet{Sobol1993SensitivityEF}. The methods implemented in \pkg{SAFE} allow one to assess the robustness and convergence of sensitivity indices. 
The R package \pkg{sensitivity} is designed to implement sensitivity analysis methods and provides the approaches for numerical calculation of the Sobol indices.  
%In particular, we can mention the implementation of the Kriging based global sensitivity analysis. 
In particular, Kriging method can be used to reduce the number of observations in global sensitivity analysis. The function \fct{sobolGP} of the package \pkg{sensitivity} builds a Kriging based meta-model using the function \fct{km} of the package \pkg{DiceKriging} \cite{JSSv051i01}, and estimates its Sobol indices. This procedure can also be done using the function \fct{km} and the function \fct{fast99} of the package \pkg{sensitivity} (see Section 4.5. of \cite{JSSv051i01}). In this case, the idea is once again to build a Kriging based meta-model using the function \fct{km} and then estimate its Sobol indices using the function \fct{fast99}. In both cases the true function is substituted by a Kriging based meta-model and then its Sobol indices are estimated. %In function \fct{sobolGP} the Sobol indices are estimated using Monte Carlo integration and in function \fct{fast99} they are estimated using extended-FAST method (\citet{doi:10.1080/00401706.1999.10485594}). 
In the sobolGP function, the Sobol indices are estimated by Monte Carlo integration, while the fast99 function estimates them using the extended-FAST method \cite{doi:10.1080/00401706.1999.10485594}.
%When dealing with large datasets and complex models, these methods may be very costly in terms of the computational time. In order to reduce the computational time, in the RKHSMetaMod package, we propose to use the empirical variances to estimate the Sobol indices (see Section \hyperref[subsec:SI]{Estimation of the Sobol indices}). This procedure is very fast and provides the estimation of the Sobol indices in a very small amount of time. 
To reduce the computational burden when dealing with large datasets and complex models, in \pkg{RKHSMetaMod} package, we propose to use the empirical variances to estimate the Sobol indices (see Section \ref{subsec:SI}).   
Besides, the estimation of the Sobol indices in the \pkg{RKHSMetaMod} package is done based on the RKHS meta-model which is a sparse estimator. It is beneficial since instead of calculating the Sobol indices of all groups $v\in\mathcal{P}$, only the Sobol indices associated with the groups in the support of the RKHS meta-model are computed (see Section \ref{subsec:SI}). Moreover, the functions \fct{sobolGP} and \fct{fast99} provide the estimation of the first order and the total Sobol indices only, while the procedure in the \pkg{RKHSMetaMod} package makes it possible to estimate the high order Sobol indices.
The R packages \pkg{DiceKriging} and \pkg{DiceOptim} (Deep Inside Computer Experiments Kriging/Optim) \cite{JSSv051i01} implement the Kriging based meta-models to estimate complex models in the high dimensional context.
They provide different GP (Kriging) models corresponding to the Gaussian, MatÃ©rn, Exponential and Power-Exponential correlation functions. 
The estimation of the parameters of the correlation functions in these packages relies on the global optimizer with gradient \fct{genoud} algorithm of the
package \pkg{rgenoud} \cite{JSSv042i11}. These packages do not implement any method of the sensitivity analysis themselves. However, some authors (see Section 4.5. of \cite{JSSv051i01} for example) perform sensitivity analysis on their estimated meta-models by employing the functions of the package \pkg{sensitivity}. 
The R package \pkg{RobustGaSP} (Robust Gaussian Stochastic Process) \cite{Gu_2019} approximates a complex model by a GP meta-model. This package implements marginal posterior mode estimation of the GP model parameters. The estimation method in this package insures the robustness of the parameter estimation in the GP model, and allows one also to identify input variables that have no effect on the variability of the function under study. 
%\item the packages which implement the meta-modelling approaches as well as the sensitivity analysis methods such as \pkg{mlegp} (\citet{mlegp}) and \pkg{UQLab} (\citet{doi:10.1061/9780784413609.257}).
%The most similar packages to the \pkg{RKHSMetaMod} are the packages which implement the meta-modelling approaches as well as the sensitivity analysis methods such as \pkg{mlegp} (\citet{mlegp}) and \pkg{UQLab} (\citet{doi:10.1061/9780784413609.257}). 
%The packages such as \pkg{mlegp} (\citet{mlegp}) and \pkg{UQLab} (\citet{doi:10.1061/9780784413609.257}), provide functions to implement both meta-modelling approaches and sensitivity analysis methods.
The R package \pkg{mlegp} (maximum likelihood estimates of Gaussian processes) \cite{mlegp} provides functions to implement both meta-modelling approaches and sensitivity analysis methods.
%The R package \pkg{mlegp} (maximum likelihood estimates of Gaussian processes) obtains maximum likelihood estimates of GP model for the output of costly computer experiments. It considers only Gaussian correlation function and first degree polynomial trend to construct GP models. The sensitivity analysis methods implemented in this package include Functional Analysis of Variance (FANOVA) decomposition, plot functions to obtain diagnostic plots, main effects, and second order interactions.
It obtains maximum likelihood estimates of GP model for the output of costly computer experiments. %It considers only Gaussian correlation function and first degree polynomial trend to construct GP models. 
The GP models are built either on the basis of the Gaussian correlation function or on the basis of the first degree polynomial trend.
The sensitivity analysis methods implemented in this package include Functional Analysis of Variance (FANOVA) decomposition, plot functions to obtain diagnostic plots, main effects, and second order interactions. 
The prediction quality of the meta-model depends on the quality of the estimation of its parameters and more precisely the estimation of parameters in the correlation functions \cite{Kennedy00bayesiancalibration}. 
The maximum likelihood estimation of these parameters often produce unstable results, and as a consequence, the obtained meta-model may have an inferior prediction quality (\cite{10.1214/17-AOS1648}, \cite{10.1214/18-BA1133}). 
%\pkg{UQLab} (Uncertainty Quantification Lab) is a Matlab toolbox for sensitivity analysis which covers many related research fields to sensitivity analysis. For example, the meta-modelling based on the polynomial Chaos expansion and GP (Kriging) modelling as well as different methods to perform  global sensitivity analysis can be done using this toolbox. 
%\end{itemize}
The \pkg{RKHSMetaMod} package is devoted to the meta-model estimation on the RKHS $\mathcal{H}$.
It implements the convex optimization algorithms to calculate meta-models; provides the functions to compute the prediction error of the obtained meta-models; performs the sensitivity analysis on the obtained meta-models and more precisely calculate their Sobol indices. 
%and performs the sensitivity analysis on the obtained meta-models by providing the estimations of their Sobol indices. 
The convex optimization algorithms used in this package are all written using C++ libraries, and are adapted to take into account the problem of high dimensionality in this context. %This package is available from the Comprehensive R Archive Network (CRAN) at \url{https://cran.r-project.org/web/packages/RKHSMetaMod/}.
This package is available from the Comprehensive R Archive Network (CRAN) \cite{rkhsmeamodpackage}.

The organization of the paper is as follows: In the next Section, we describe the estimation method. In Section \ref{sec:Optalgo}, we present in details the algorithms used in the \pkg{RKHSMetaMod} package to obtain the RKHS meta-model. In Section \ref{sec:Overview}, an overview of the package functions as well as a brief documentation of them are given. 
Section \ref{sec:examples} includes two parts: In the first part, Section \ref{subsec:simulstdy}, the performance of the \pkg{RKHSMetaMod} package functions is validated through a simulation study. In the second part, Section \ref{subsec:comprexpl}, the comparison in terms of the predictive accuracy between the RKHS meta-model and the Kriging based meta-models from \pkg{RobustGaSP} \cite{Gu_2019} and \pkg{DiceKriging} \cite{JSSv051i01} packages is given through two examples.
\section{Estimation method}\label{sec:Description}
In this Section, we present: the RKHS ridge group sparse and the RKHS group lasso procedures (see \ref{subsec:RGS}), %the method of \citet{DURRANDE201357} to construct the RKHS $\mathcal{H}$ (see \hyperref[subsec:Data]{RKHS construction}), 
the strategy of choosing the tuning parameters in the RKHS ridge group sparse algorithm (see \ref{subsec:Err}), and the calculation of the empirical Sobol indices of the RKHS meta-model (see \ref{subsec:SI}).
\subsection{RKHS ridge group sparse and RKHS group lasso procedures}\label{subsec:RGS}
Let us denote by $n$ the number of observations. The dataset consists of a vector of $n$ observations $Y=(Y_1,...,Y_n)$, and a $n\times d$ matrix of features $X$ with components $(X_{ai},i=1,...,n,a=1,...,d)\in\mathbb{R}^{n\times d}.$ 
For some tuning parameters $\gamma_v$, $\mu_v$, $v\in\mathcal{P}$, the RKHS ridge group sparse criterion is defined by, 
\begin{align}
\label{functional}
\mathcal{L}(f)=\frac{1}{n}\sum_{i=1}^n\Big(Y_i-f_0-\sum_{v\in\mathcal{P}}f_v(X_{vi}) \Big)^2
+\sum_{v\in\mathcal{P}}\gamma_v\Vert f_v\Vert_n+\sum_{v\in\mathcal{P}}\mu_v\Vert f_v\Vert_{\mathcal{H}_v} ,
\end{align}
where $X_v$ represents the matrix of variables corresponding to the $v$-th group, i.e. $X_v=(X_{vi},i=1,...,n,v\in\mathcal{P})\in\mathbb{R}^{n\times \vert \mathcal{P}\vert},$ 
and where $\Vert f_v\Vert_n$ is the empirical $L^2$-norm of $f_v$ defined by the sample $\{X_{vi}\}_{i=1}^n$ as $\Vert f_v\Vert_n=\sqrt{\sum_{i=1}^nf_v^2(X_{vi})/n}.$

The penalty function in the criterion \eref{functional} is the sum of the Hilbert norm and the empirical norm, which allows us to select few terms in the additive decomposition of $f$ over sets $v \in \mathcal{P}$. Moreover, the Hilbert norm favours the smoothness of the estimated $f_v$, $v \in \mathcal{P}$.\\ 
Let $\cF = \{ f:\: f=f_{0} + \sum_{v\in \cP} f_{v},\: f_v \in \cH_{v},\: \|f_v\|_{\cH_v} \leq r_v,\:r_v\in\mathbb{R}^{+} \}$ be the set of functions.
Then the RKHS meta-model is defined by,
\begin{equation}
\label{ch3prediction}
\widehat{f}=\argmin_{f\in\mathcal{F}}\mathcal{L}(f).
\end{equation}
According to the Representer Theorem \cite{kimeldorf1970}, the non-parametric functional minimization problem described above is equivalent to a parametric minimization problem. Indeed, the solution of the minimization problem \eref{ch3prediction} belonging to the RKHS $\mathcal{H}$ is written as $f=f_0+\sum_{v\in\mathcal{P}}f_v$, where for some matrix $\theta=(\theta_{vi},i=1,...,n,v\in\mathcal{P})\in\mathbb{R}^{n\times\vert\mathcal{P}\vert}$ we have for all $v\in\mathcal{P}$, 
\begin{align}
\label{normhvdef}
f_v(.)=\sum_{i=1}^n\theta_{vi}k_v(X_{vi},.),\mbox{ and }\Vert f_v\Vert^2_{\mathcal{H}_v}=\sum_{i,i'=1}^n\theta_{vi}\theta_{vi'}k_v(X_{vi},X_{vi'}).
\end{align}
Let $\Vert.\Vert$ be the Euclidean norm (called also $L^2$-norm) in $\mathbb{R}^n$, and for each $v\in\mathcal{P}$, let $K_v$ be the $n\times n$ Gram matrix associated with the kernel $k_v(.,.)$, i.e. $(K_v)_{i,i'}=k_v(X_{vi},X_{vi'})$. Let also $K^{1/2}$ be the matrix that satisfies $t(K^{1/2})K^{1/2}=K$, and let $\widehat{f}_0$ and $\widehat{\theta}$ be the minimizers of the following penalized least-squares criterion:
\begin{align*}
C(f_0,\theta)=\Vert Y-f_0 I_n-\sum_{v\in\mathcal{P}}K_v\theta_v\Vert^2+\sqrt{n}\sum_{v\in\mathcal{P}}\gamma_v\Vert K_v\theta_v\Vert+n\sum_{v\in\mathcal{P}}\mu_v\Vert K_v^{1/2}\theta_v\Vert.
\end{align*}
Then the estimator $\widehat{f}$ defined in Equation \eref{ch3prediction} satisfies, %$\widehat{f}(X)=\widehat{f}_0+\sum_{v\in\mathcal{P}}\widehat{f}_v(X_v)$ where $\widehat{f}_v(X_v)=\sum_{i=1}^n\widehat{\theta}_{vi}k_v(X_{vi},X_v).$
\begin{align*}
\widehat{f}(X)=\widehat{f}_0+\sum_{v\in\mathcal{P}}\widehat{f}_v(X_v)\mbox{ with }\widehat{f}_v(X_v)=\sum_{i=1}^n\widehat{\theta}_{vi}k_v(X_{vi},X_v).
\end{align*}
\begin{rem}  The constraint $\Vert f_v\Vert_{\mathcal{H}_v}\leq r_v$ %is not taken into account in the parametric minimization problem. This constraint 
is crucial for theoretical properties, but the value of $r_v$ is generally unknown and has no practical usefulness. In this package, it is not taken into account in the parametric minimization problem.  
\end{rem}
For each $v\in\mathcal{P}$, let $\gamma'_v$ and $\mu'_v$ be the weights that are chosen suitably. We define $\gamma_v=\gamma\times\gamma'_v$ and $\mu_v=\mu\times\mu'_v$ with $\gamma,\mu\in\mathbb{R}^+$.
%\begin{align*}
%\gamma_v=\gamma\times\gamma'_v\mbox{ and }\mu_v=\mu\times\mu'_v\mbox{ with } \gamma,\mu\in\mathbb{R}^+.
%\end{align*} 
\begin{rem} This formulation simplifies the choice of the tuning parameters since instead of tuning $2\times\vert\mathcal{P}\vert$ parameters $\gamma_v$ and $\mu_v$, $v\in\mathcal{P}$, only two parameters $\gamma$ and $\mu$ are tuned. Moreover, the weights $\gamma'_v$ and $\mu'_v$, $v\in\mathcal{P}$ may be of interest in practice. For example, one can take weights that increase with the cardinal of $v$ in order to favour the effects with small interaction order between variables.
\end{rem}
For the sake of simplicity, in the rest of this paper for all $v\in\mathcal{P}$ the weights $\gamma'_v$ and $\mu'_v$ are assumed to be set as one, and the RKHS ridge group sparse criterion is then expressed as follows:
\begin{align}
\label{parametric}
C(f_0,\theta)=\Vert Y-f_0 I_n-\sum_{v\in\mathcal{P}}K_v\theta_v\Vert^2+\sqrt{n}\gamma\sum_{v\in\mathcal{P}}\Vert K_v\theta_v\Vert+n\mu\sum_{v\in\mathcal{P}}\Vert K_v^{1/2}\theta_v\Vert.
\end{align}
%By considering only the second part of the penalty function in the RKHS ridge group sparse criterion \eref{parametric}, i.e. by setting $\gamma=0$, the RKHS group lasso criterion is obtained as follows:
If we consider only the second part of the penalty function in the criterion \eref{parametric} ( i.e. set $\gamma=0$), we obtain the RKHS group lasso criterion as follows:
\begin{align}
\label{Lasso}
C_g(f_0,\theta)=\Vert Y-f_0I_n-\sum_{v\in\mathcal{P}}K_v\theta_v\Vert^2+n\mu\sum_{v\in\mathcal{P}}\Vert K_v^{1/2}\theta_v\Vert,
\end{align}
which is a group lasso criterion \cite{doi:10.1111/j.1467-9868.2005.00532.x} up to a scale transformation.

In the \pkg{RKHSMetaMod} package, the RKHS ridge group sparse algorithm is initialized using the solutions obtained by solving the RKHS group lasso algorithm.
Indeed, the penalty function in the RKHS group lasso criterion \eref{Lasso} insures the sparsity in the solution. Therefore, for a given value of $\mu$, by implementing the RKHS group lasso algorithm (see Section \ref{subsec:optimGL}), a RKHS meta-model with few terms in its additive decomposition is obtained. The support and the coefficients of a RKHS meta-model which is obtained by implementing RKHS group lasso algorithm will be denoted by $\widehat{S}_{\widehat{f}_{\text{Group Lasso}}}$ and $\widehat{\theta}_{\text{Group Lasso}}$, respectively.
From now on the tuning parameter in the RKHS group lasso criterion will be denoted by:
\begin{equation}
\label{tungrp}
\mu_g=\sqrt{n}\mu.
\end{equation}
\subsection{Choice of the tuning parameters}\label{subsec:Err} 
While dealing with an optimization problem of a criterion of the form (\ref{parametric}), one of the essential steps is to choose appropriately the tuning parameters. Cross-validation is generally used for that purpose.
Nevertheless in the context of high-dimensional complex models, the
computational time for a cross-validation procedure may be prohibitively
high. Therefore we propose a procedure based on a single testing data set:
\begin{itemize}
\item we first choose, a grid of values of the tuning parameters $\mu$ and $\gamma$;\\
Let $\mu_{\text{max}}$ be the smallest value of $\mu_g$ (see Equation \eref{tungrp}), such that the solution to the minimization of the RKHS group lasso problem for all $v\in\mathcal{P}$ is $\theta_v=0$. We have, 
\begin{align}
\label{maxmu}
\mu_{\text{max}}=\max_v\Big(2\Vert K_v^{1/2}(Y-\bar{Y})\Vert\Big)/\sqrt{n}.
\end{align}
In order to set up the grid of values of $\mu$, one may find $\mu_{\text{max}}$ and then a grid of values of $\mu$ could be defined by %follows:
%$$\mu_l=\frac{\mu_{\text{max}}}{(\sqrt{n}\times2^{l})},\: l\in\{1,...,l_{\text{max}}\}.$$ 
$\mu_l=\mu_{\text{max}}/(\sqrt{n}\times2^{l})$ for $l\in\{1,...,l_{\text{max}}\}.$
The grid of values of $\gamma$ is chosen by the user.  

\item next, for the grid of values of $\mu$ and $\gamma$, we calculate a sequence of estimators. Each estimator associated with the pair $(\mu,\gamma)$ in the grid of values of $\mu$ and $\gamma$, denoted by $\widehat{f}_{(\mu,\gamma)}$, is the solution of the RKHS ridge group sparse optimization problem or the RKHS group lasso optimization problem if $\gamma=0$. 
\item finally, the obtained estimators $\widehat{f}_{(\mu,\gamma)}$ are evaluated using a testing dataset, 
%$$\{(Y^{\text{test}}_i,X^{\text{test}}_i)\}_{i=1}^{n^{\text{test}}}.$$ 
$\{(Y^{\text{test}}_i,X^{\text{test}}_i)\}_{i=1}^{n^{\text{test}}}.$
The prediction error associated with the estimator $\widehat{f}_{(\mu,\gamma)}$ is calculated by,
%\begin{align*}
%\text{ErrPred}(\mu,\gamma)=\frac{1}{n^{\text{test}}}\sum_{i=1}^{n^{\text{test}}}(Y^{\text{test}}_i-\widehat{f}_{(\mu,\gamma)}(X^{\text{test}}_i))^2,
%\end{align*}
$$\text{ErrPred}(\mu,\gamma)=\sum_{i=1}^{n^{\text{test}}}(Y^{\text{test}}_i-\widehat{f}_{(\mu,\gamma)}(X^{\text{test}}_i))^2/n^{\text{test}},$$
where for $S_{\widehat{f}}$ being the support of the estimator $\widehat{f}_{(\mu,\gamma)}$ we have, 
$$\widehat{f}_{(\mu,\gamma)}(X^{\text{test}})=\widehat{f}_0+\sum_{v\in S_{\widehat{f}}}\sum_{i=1}^n\widehat{\theta}_{vi}k_v(X_{vi},X^{\text{test}}_v).$$
The pair $(\widehat{\mu},\widehat{\gamma})$ with the smallest value of the prediction error is chosen, and the estimator $\widehat{f}_{(\widehat{\mu},\widehat{\gamma})}$ is considered as the \textit{best} estimator of the function $m$, in terms of the prediction error.
\end{itemize}

In the \pkg{RKHSMetaMod} package, the algorithm to calculate a sequence of the RKHS meta-models, the value of $\mu_{\text{max}}$, and the prediction error are implemented as \fct{RKHSMetMod}, \fct{mu$\_$max}, and \fct{PredErr} functions, respectively. These functions are described in Section \ref{sec:Overview}, and illustrated in Example \ref{ex:1}, Example \ref{ex:3}, and Examples \ref{ex:1}, \ref{ex:3}, \ref{ex:4}, respectively.
\subsection{Estimation of the Sobol indices}\label{subsec:SI}
The variance of the function $m$ is estimated by the variance of the estimator $\widehat{f}$. As the estimator $\widehat{f}$ belongs to the RKHS $\mathcal{H}$, it admits the Hoeffding decomposition and, %$\text{var}(\widehat{f}(X))=\sum_{v\in\mathcal{P}}\text{var}(\widehat{f}_v(X_v)),$
\begin{align*}
\text{var}(\widehat{f}(X))=\sum_{v\in\mathcal{P}}\text{var}(\widehat{f}_v(X_v)),\mbox{ where }\forall v\in\mathcal{P},\: \text{var}(\widehat{f}_v(X_v))=E_X({\widehat{f}_v}^2(X_v))=\Vert \widehat{f}_v\Vert_2^2.
\end{align*}
%where for all $v\in\mathcal{P}$, $\text{var}(\widehat{f}_v(X_v))=E_X({\widehat{f}_v}^2(X_v))=\Vert \widehat{f}_v\Vert_2^2.$
%\begin{align*}
%\text{var}(\widehat{f}_v(X_v))=E_X({\widehat{f}_v}^2(X_v))=\Vert \widehat{f}_v\Vert_2^2.
%\end{align*}
In order to reduce the computational cost in practice, one may estimate the variances of $\widehat{f}_v(X_v)$, $v\in\mathcal{P}$ by their empirical variances. Let $\widehat{f}_{v.}$ be the empirical mean of $\widehat{f}_v(X_{vi})$, $i=1,...,n$, then: %$\widehat{\text{var}}(\widehat{f}_v(X_v))=\sum_{i=1}^n(\widehat{f}_v(X_{vi})-\widehat{f}_{v.})^2/(n-1).$
\begin{align*}
\widehat{\text{var}}(\widehat{f}_v(X_v))=\frac{1}{n-1}\sum_{i=1}^n(\widehat{f}_v(X_{vi})-\widehat{f}_{v.})^2.
\end{align*}
%For the groups $v$ that belong to the support of $\widehat{f}$, the estimators of the Sobol indices of $m$ are defined by, $\widehat{S}_v=\widehat{\text{var}}(\widehat{f}_v(X_v))/\sum_{v\in\mathcal{P}}\widehat{\text{var}}(\widehat{f}_v(X_v)),$
%\begin{align*}
%\widehat{S}_v=\frac{\widehat{\text{var}}(\widehat{f}_v(X_v))}{\sum_{v\in\mathcal{P}}\widehat{\text{var}}(\widehat{f}_v(X_v))},
%\end{align*}
%and for the groups $v$ that do not belong to the support of $\widehat{f}$, we have $\widehat{S}_v=0$.
For the groups $v$ that do not belong to the support of $\widehat{f}$, we have $\widehat{S}_v=0$ and for the groups $v$ that belong to the support of $\widehat{f}$, the estimators of the Sobol indices of $m$ are defined by,
\begin{align*}
\widehat{S}_v=\widehat{\text{var}}(\widehat{f}_v(X_v))/\sum_{v\in\mathcal{P}}\widehat{\text{var}}(\widehat{f}_v(X_v)).
\end{align*}
In the \pkg{RKHSMetaMod} package, the algorithm to calculate the empirical Sobol indices $\widehat{S}_v$, $v\in\mathcal{P}$ is implemented as \fct{SI$\_$emp} function. This function is described in Section \ref{subsec:Comp} and illustrated in Examples \ref{ex:1}, \ref{ex:3}, \ref{ex:4}.
\section{Algorithms}\label{sec:Optalgo}
The \pkg{RKHSMetaMod} package implements two optimization algorithms: the RKHS ridge group sparse (see Algorithm \ref{algo:RGS}) and the RKHS group lasso (see Algorithm \ref{algo:GL}). These algorithms rely on the Gram matrices $K_v$, $v\in\mathcal{P}$ that have to be positive definite. Therefore, the first and essential step in this package is to calculate these matrices and insure their positive definiteness. This step is detailed in an algorithm that is described in the next Section. The second step is to estimate the RKHS meta-model. In the \pkg{RKHSMetaMod} package, two different objectives based on different procedures are considered to calculate this estimator:
\begin{itemize}
\item[1.] The RKHS meta-model with the \textit{best} prediction quality.

The procedure to calculate the RKHS meta-model with the \textit{best} prediction quality has been described in Section \ref{subsec:Err}: a sequence of values of the tuning parameters $(\mu,\gamma)$ is considered, and the RKHS meta-models associated with each pair of the values of $(\mu,\gamma )$ are calculated. For $\gamma=0$, the RKHS meta-model is obtained by solving the RKHS group lasso optimization problem, while for $\gamma\neq 0$ the RKHS ridge group sparse optimization problem is solved to calculate the RKHS meta-model. The obtained RKHS meta-models are evaluated by considering a new dataset and the RKHS meta-model with the minimum value of the prediction error is chosen as the \textit{best} estimator.
\item[2.] The RKHS meta-model with at most $qmax$ groups in its support, i.e. $\vert S_{\widehat{f}}\vert\leq qmax$.
%The RKHS meta-model with at most $qmax$ active groups. 

%By active groups we mean the groups $v\in\mathcal{P}$ that are not zero and constitute then the support $S_{\widehat{f}}$ of the obtained meta-model $\widehat{f}$. Therefore, the RKHS meta-model with at most $qmax$ active groups has at most $qmax$ groups in its support, i.e. $\vert S_{\widehat{f}}\vert\leq qmax$. The procedure to calculate such a meta-model is detailed in Section \hyperref[subsec:optimqmax]{RKHS meta-model with $qmax$ active groups}: first, the tuning parameter $\gamma$ is set as zero. Then, a value of $\mu$ for which the number of groups $v\in\mathcal{P}$ in the solution of the RKHS group lasso optimization problem is equal to $qmax$, is computed. This value will be denoted by $\mu_{qmax}$. Finally, the RKHS meta-models with at most $qmax$ active groups are obtained by implementing the RKHS ridge group sparse algorithm for a grid of values of $\gamma\neq0$ and $\mu_{qmax}$. 
First, the tuning parameter $\gamma$ is set as zero. Then, a value of $\mu$ for which the number of groups $v\in\mathcal{P}$ in the solution of the RKHS group lasso optimization problem is equal to $qmax$, is computed. This value of $\mu$ will be denoted by $\mu_{qmax}$. Finally, the RKHS meta-models containing at most $qmax$ groups in their support are obtained by implementing the RKHS ridge group sparse algorithm for a grid of values of $\gamma\neq0$ and $\mu_{qmax}$. This procedure is described in more details in Section \ref{subsec:optimqmax}.
\end{itemize}

\subsection{Calculation of the Gram matrices}\label{subsec:Gramm}
The available kernels in the \pkg{RKHSMetaMod} package are: Gaussian kernel, Matérn $3/2$ kernel, Brownian kernel, quadratic kernel and linear kernel. The usual presentation of these kernels is given in Table \ref{kernels}. 
\begin{table}[h!]
\centering
\begin{tabular}{l|l|l} 
 Kernel type &  Mathematics formula for $u\in\mathbb{R}^n,v\in\mathbb{R}$ & RKHSMetaMod name \\ \hline
Gaussian & $k_a(u,v)=\exp(-\Vert u-v\Vert^2/2r^2)$ & "gaussian"\\ 
Matérn 3/2 & $k_a(u,v)=(1+\sqrt{3}\vert u-v\vert/r)\exp(-\sqrt{3}\vert u-v\vert/r)$ & "matern"\\
Brownian & $k_a(u,v)=\min(u,v)+1$ & "brownian"\\
Quadratic & $k_a(u,v)=(u^Tv+1)^2$ & "quad"\\
Linear & $k_a(u,v)=u^Tv+1$ &"linear"\\  
\end{tabular}
\caption{List of the reproducing kernels used to construct the RKHS $\mathcal{H}$. The range parameters $r$ in the Gaussian and Matérn $3/2$ kernels are assumed to be fixed and set as $1/2$ and $\sqrt{3}/2$, respectively. The value $1$ is added to the Brownian kernel to relax the constraint of nullity at the origin.\label{kernels}} %\cite{DURRANDE201357}. \label{kernels}}
\end{table}
The choice of the kernel that is done by the user determines the functional approximation space. For a chosen kernel, the algorithm to calculate the Gram matrices $K_v$, $v\in\mathcal{P}$ in the \pkg{RKHSMetaMod} package, is implemented as \fct{calc$\_$Kv} function. This algorithm is based on three essential points:
\begin{itemize}
\item[(1)] Modify the chosen kernel:\\ 
In order to satisfy the conditions of constructing the RKHS $\mathcal{H}$ described in Section \ref{subsec:Data}, these kernels are modified according to Equation \eref{kernelmodify}. Let us take the example of the Brownian kernel:\\
%\begin{exemp}
The RKHS associated with the Brownian kernel $k_a(X_a,X_a')=\min(X_a,X_a')+1$ is well known to be %the set%,
%\begin{align*}
%\mathcal{H}_a=\Big\{f:[0,1]\rightarrow \mathbb{R} \mbox{ is absolutely continuous, and }f(0)=0,\:\int_0^1{f'(X_a)}^2dX_a<\infty\Big\},
%\end{align*}
%with the inner product
%\begin{align*}
%\langle f,h\rangle_{\mathcal{H}_a}=\int_0^1f'(X_a)h'(X_a)dX_a.
%\end{align*}
$\mathcal{H}_a=\{f:[0,1]\rightarrow \mathbb{R} \mbox{ is absolutely continuous, and }f(0)=0,\:\int_0^1{f'(X_a)}^2dX_a<\infty\},$
with the inner product $\langle f,h\rangle_{\mathcal{H}_a}=\int_0^1f'(X_a)h'(X_a)dX_a.$ %The kernel $k_{0a}$ associated with the Brownian kernel is calculated as follows,
Easy calculations lead to obtain the Brownian kernel as follows,
%\begin{align*}
%k_{0a}&=\min (X_a,X_a')+1-\frac{(\int_0^1(\min(X_a,U)+1)dU)(\int_0^1(\min(X_a',U)+1)dU)}{(\int_0^1\int_0^1(\min(U,V)+1)dUdV)},\\
%&=\min(X_a,X_a')+1-\frac{3}{4}(1+X_a-\frac{X_a^2}{2})(1+X_a'-\frac{X_a'^2}{2}).
%\end{align*}
\begin{align*}
k_{0a}=\min(X_a,X_a')+1-(3/4)(1+X_a-{X_a^2/2})(1+X_a'-{X_a'^2/2}).
\end{align*}
%The RKHS associated with the kernel $k_{0a}$ is the set, 
%\begin{align*}
%\mathcal{H}_{0a}=\Big\{f\in\mathcal{H}_a:\:\int_{0}^1f(X_a)dX_a=0\Big\}.
%\end{align*}
The RKHS associated with the kernel $k_{0a}$ is the set $\mathcal{H}_{0a}=\{f\in\mathcal{H}_a:\:\int_{0}^1f(X_a)dX_a=0\}$, and we have $\mathcal{H}= \mathbbm{1} + \sum_{v \in \mathcal{P}} \mathcal{H}_{v}=\{f:[0,1]^d\rightarrow\mathbb{R}:\:f=f_0+\sum_{v\in\mathcal{P}}f_v(X_v),\mbox{ with }f_v\in\mathcal{H}_v\}.$ 
%Finally, the RKHS $\mathcal{H}= \mathbbm{1} + \sum_{v \in \mathcal{P}} \mathcal{H}_{v}$
% is the following set,
%\begin{align*}
%\mathcal{H}=\Big\{f:[0,1]^d\rightarrow\mathbb{R}:\:f=f_0+\sum_{v\in\mathcal{P}}f_v(X_v),\mbox{ with }f_v\in\mathcal{H}_v\Big\}.
%\end{align*}
%\end{exemp}
\begin{rem}\label{ch3:remxun}
In the current version of the package, we consider the input variables $X=(X_1,...,X_d)$ that are uniformly distributed on $[0,1]^d$. In order to consider the input variables that are not distributed uniformly, it suffices to modify a part of the function \fct{calc$\_$Kv} related to the calculation of the kernels $k_{0a}$, $a=1,...,d$. For example, for $X=(X_1,...,X_d)$ being distributed with law $P_X=\prod_{a=1}^d P_a$ on $\mathcal{X}=\bigotimes_{a=1}^d\mathcal{X}_a\subset\mathbb{R}^d$, the kernel $k_{0a}$ associated with the Brownian kernel is calculated as follows, 
\begin{align*}
k_{0a}&=\min (X_a,X_a')+1-\frac{(\int_{\mathcal{X}_a}(\min(X_a,U)+1)dP_a)(\int_{\mathcal{X}_a}(\min(X_a',U)+1)dP_a)}{(\int_{\mathcal{X}_a}\int_{\mathcal{X}_a}(\min(U,V)+1)dP_adP_a)}.
\end{align*}
The other parts of the function \fct{calc$\_$Kv} remain unchanged.
\end{rem}
\item[(2)] Calculate the Gram matrices $K_v$ for all $v$:\\
First, for all $a=1,...,d$, the Gram matrices $K_a$ associated with kernels $k_{0a}$ are calculated using Equation \eref{kernelmodify}, $(K_a)_{i,i'}=k_{0a}(X_{ai},X_{ai'}).$
%$$(K_a)_{i,i'}=k_{0a}(X_{ai},X_{ai'}).$$
%For all $a=1,...,d$, the Gram matrices $K_a$ associated with kernels $k_{0a}$ are calculated using Equation \eref{kernelmodify}, $(K_a)_{i,i'}=k_{0a}(X_{ai},X_{ai'}).$
Then, for all $v\in\mathcal{P}$, the Gram matrices $K_v$ associated with kernel $k_v=\prod_{a\in v}k_{0a}$ are %calculated as follows: 
%$$K_v=\bigodot_{a\in v} K_a,$$
computed by $K_v=\bigodot_{a\in v} K_a,$
%For groups $v\in\mathcal{P}$, $\vert v\vert\geq 2$ the Gram matrices $K_v$ associated with kernel $k_v=\prod_{a\in v}k_{0a}$ are computed by $K_v=\bigodot_{a\in v} K_a,$ where $\bigodot$ denotes the Hadamard product.
\item[(3)] Insure the positive definiteness of the matrices $K_v$:\\
The output of the function \fct{calc$\_$Kv} is one of the input arguments of the functions associated with the RKHS group lasso and the RKHS ridge group sparse algorithms. Throughout these algorithms we need to calculate the inverse and the square root of the matrices $K_v$. In order to avoid the numerical problems and insure the invertibility of the matrices $K_v$, it is mandatory to have these matrices positive definite. One way to render the matrices $K_v$ positive definite is to add a nugget effect to them. That is, to modify matrices $K_v$ by adding a diagonal with a constant term, i.e. $K_v+\mbox{epsilon}\times I_n$. The value of $\mbox{epsilon}$ is computed based on the data and through a part of the algorithm of the function \fct{calc$\_$kv}. %The options "correction" and "tol" are provided by the function \fct{calc$\_$Kv} in order to activate this part of its algorithm leading to modify the matrices $K_v$, $v\in\mathcal{P}$ and render them positive definite if necessary.
% This part of the algorithm can be activated via the arguments, "correction" and "tol", of the function \fct{calc$\_$Kv}.  
%The options, "correction" and "tol", are provided by the function \fct{calc$\_$Kv} in order to calculate the value of $\mbox{epsilon}$ and modify the matrices $K_v$, $v\in\mathcal{P}$ to render them positive definite if necessary.
%Throughout these algorithms we need to calculate the inverse and the square root of the matrices $K_v$. In order to avoid the numerical problems and insure the invertibility of the matrices $K_v$, one possibility is to render them positive definite by adding a nugget effect. That is to modify matrices $K_v$ by adding a diagonal with a constant term, $K_v+\mbox{epsilon}\times I_n$. The options, "correction" and "tol", are provided by the function \fct{calc$\_$Kv} in order to calculate the value of $\mbox{epsilon}$ and modify the matrices $K_v$, $v\in\mathcal{P}$ to render them positive definite if necessary.
%The value of $\eta$ is estimated based on the data and using the following algorithm: 
%As both of these algorithms rely on the positive definiteness of these matrices, it is mandatory to have $K_v$, $v\in\mathcal{P}$ that are positive definite. The options, "correction" and "tol", are provided by the function \fct{calc$\_$Kv} in order to insure the positive definiteness of the matrices $K_v$, $v\in\mathcal{P}$. 
Let us briefly explain this part of the algorithm:\\
For each group $v\in\mathcal{P}$, let $\lambda_{v,i},i=1,...,n$ be the eigenvalues associated with the matrix $K_v$. Set $\lambda_{v,\text{max}}={\text{max}}_{i}\lambda_{v,i}$ and $\lambda_{v,\text{min}}={\text{min}}_{i}\lambda_{v,i}$. For some fixed value of tolerance "tol", and for each matrix $K_v$, if $\lambda_{v,\text{min}} < \lambda_{v,\text{max}}\times\text{tol}$:   
%$$\text{"if } \lambda_{v,\text{min}} < \lambda_{v,\text{max}}\times\text{tol"},$$ 
%then the "correction" to $K_v$ is done. That is, 
%$$\text{"The eigenvalues of } K_v \text{ are replaced by }\lambda_{v,i}+\text{epsilon"},$$ 
then, the eigenvalues of $K_v$ are replaced by $\lambda_{v,i}+\text{epsilon}$, with "epsilon" being equal to $\lambda_{v,\text{max}}\times$"tol".
The value of "tol" is set as $1e^{-8}$ by default, but one may consider a smaller or a greater value for it depending on the kernel chosen and the value of $n$. 
\end{itemize}
The function \fct{calc$\_$Kv} is described in Section \ref{subsec:Comp} and illustrated in Example \ref{ex:3}.
\subsection{Optimization algorithms}\label{subsec:optim}
The RKHS meta-model is the solution of one of the optimization problems: the minimization of the RKHS group lasso criterion presented in Equation \eref{Lasso} (if $\gamma=0$), or the minimization of the RKHS ridge group sparse criterion presented in Equation \eref{parametric} (if $\gamma\neq0$). 
In the following, the algorithms to solve these optimization problems are presented.
\subsubsection{RKHS group lasso}\label{subsec:optimGL}
A popular technique for doing group wise variable selection is group lasso. With this procedure, depending on the value of the tuning parameter $\mu$, an entire group of predictors may drop out of the model. An efficient algorithm for solving group lasso problem is the classical block coordinate descent algorithm (\cite{10.1561/2200000016}, \cite{10.1561/2200000050}). Following the idea of \cite{10.2307/1390712}, \cite{doi:10.1111/j.1467-9868.2005.00532.x} implemented a block wise descent algorithm for the group lasso penalized least-squares under the condition that the model matrices in each group are orthonormal. A block coordinate (gradient) descent algorithm for solving the group lasso penalized logistic regression is then developed by \cite{MeiGeeBue08}. This algorithm is implemented in the R package \pkg{grplasso} available from CRAN \cite{grppkg}. \cite{Yang:2015:FUA:2833490.2833520} proposed a unified algorithm named group wise majorization descent for solving the general group lasso learning problems by assuming that the loss function satisfies a quadratic majorization condition. The implementation of their work is done in the \pkg{gglasso} R package available from CRAN \cite{gglspkg}. 

In order to solve the RKHS group lasso optimization problem, we use the classical block coordinate descent algorithm. The minimization of criterion $C_g(f_0,\theta)$ (see Equation \eref{Lasso}) is done along each group $v$ at a time. At each step of the algorithm, the criterion $C_g(f_0,\theta)$ is minimized as a function of the current block's parameters, while the parameters values for the other blocks are fixed to their current values. The procedure is repeated until convergence. This procedure leads to Algorithm \ref{algo:GL} (see \ref{app:technical} for more details on this procedure).
\begin{algorithm}[h!]%[htbp]%[h!]
\caption{RKHS group lasso algorithm:}\label{algo:GL}
\begin{algorithmic}[1]
\State{Set $\theta_0=[0]_{\vert\mathcal{P}\vert\times n}$}
\Repeat
 \State{Calculate $f_0=\text{argmin}_{f_0}C_g(f_0,\theta)$}
 \For{$v\in \mathcal{P}$}
  \State{Calculate $R_v=Y-f_0-\sum_{v\neq w}K_w\theta_w$}
   \If{$\Vert\frac{2}{\sqrt{n}} K_v^{1/2}R_v\Vert\leq \mu_g$}
      \State $\theta_v\gets 0$
    \Else
      \State$\theta_v\gets\text{argmin}_{\theta_v}C_g(f_0,\theta)$
    \EndIf 
 \EndFor
\Until{convergence}
\end{algorithmic}
\end{algorithm}
In the \pkg{RKHSMetaMod} package, the Algorithm \ref{algo:GL} is implemented as \fct{RKHSgrplasso} function. This function is described in Section \ref{subsec:Comp}.
\subsubsection{RKHS ridge group sparse}\label{subsec:optimRGS} 
In order to solve the RKHS ridge group sparse optimization problem, we propose an adapted block coordinate descent algorithm. This algorithm is provided in two steps:
\begin{itemize}
\item[\textbf{Step $1$}]\label{step1} Initialize the input parameters by the solutions of the RKHS group lasso algorithm for each value of the tuning parameter $\mu$, and implement the RKHS ridge group sparse algorithm through the active support of the RKHS group lasso solutions until it achieves convergence.
This step is provided in order to decrease the execution time. In fact, instead of implementing the RKHS ridge group sparse algorithm over the set of all groups $\mathcal{P}$, it is implemented only over the groups in the support of the solution of the RKHS group lasso algorithm, $\widehat{S}_{\widehat{f}_{\text{Group Lasso}}}$. 
\item[\textbf{Step $2$}]\label{step2} Re-initialize the input parameters with the obtained solutions of \hyperref[step1]{Step $1$} and implement the RKHS ridge group sparse algorithm through all groups in $\mathcal{P}$ until it achieves convergence. 
This second step makes it possible to verify that no group is missing in the output of \hyperref[step1]{Step $1$}.
\end{itemize}
This procedure leads to Algorithm \ref{algo:RGS} (see \ref{app:technical} for more details on this procedure). 
\begin{algorithm}[h!]%[htbp]%[h!]
\caption{RKHS ridge group sparse algorithm:}\label{algo:RGS}
\begin{algorithmic}[1]
\State{\textbf{Step 1:}}
\State{Set $\theta_0=\widehat{\theta}_{\text{Group Lasso}}$ and $\widehat{\mathcal{P}}=\widehat{S}_{\widehat{f}_{\text{Group Lasso}}}$}
\Repeat
 \State{Calculate $f_0=\text{argmin}_{f_0}C(f_0,\theta)$}
 \For{$v\in \widehat{\mathcal{P}}$}
  \State{Calculate $R_v=Y-f_0-\sum_{v\neq w}K_w\theta_w$}
   \State{Solve $J^*=\text{argmin}_{\widehat{t}_v\in\mathbb{R}^n}\{J(\widehat{t}_v),\text{ such that }\Vert K_v^{-1/2}\widehat{t}_v\Vert\leq 1\}$}
    \If{$J^*\leq \gamma$}
      \State $\theta_v\gets 0$
    \Else
      \State$\theta_v\gets\text{argmin}_{\theta_v}C(f_0,\theta)$
    \EndIf 
 \EndFor
\Until{convergence}

\State{\textbf{Step 2:}}
\State{Implement the same procedure as \textbf{Step 1} with $\theta_0=\widehat{\theta}_{\text{old}}$, $\widehat{\mathcal{P}}=\mathcal{P}$}\Comment{$\widehat{\theta}_{\text{old}}$ is the estimation of $\theta$ in \textbf{Step 1}.}
\end{algorithmic}
\end{algorithm}
In the \pkg{RKHSMetaMod} package the Algorithm \ref{algo:RGS} is implemented as \fct{pen$\_$MetMod} function. This function is described in Section \ref{subsec:Comp} and illustrated in Example \ref{ex:3}.
\subsubsection{RKHS meta-model with at most $qmax$ groups in its support}\label{subsec:optimqmax}
By considering some prior information about the data, one may be interested in a RKHS meta-model $\widehat{f}$ with the number of groups in its support not greater than some "$qmax$". In order to obtain such an estimator, we provide the following procedure in the \pkg{RKHSMetaMod} package: 
\begin{itemize}
\item First, the tuning parameter $\gamma$ is set as zero and a value of $\mu$ for which the solution of the RKHS group lasso algorithm, Algorithm \ref{algo:GL}, contains exactly $qmax$ groups in its support is computed. This value is denoted by $\mu_{qmax}$.
\item Then, the RKHS ridge group sparse algorithm, Algorithm \ref{algo:RGS}, is implemented by setting the tuning parameter $\mu$ equal to $\mu_{qmax}$ and a grid of values of the tuning parameter $\gamma>0$.
\end{itemize}
This procedure leads to Algorithm \ref{algo:GLqmax}. 
\begin{algorithm}[h!]%[htbp]%[h!]
\caption{Algorithm to estimate RKHS meta-model with at most $qmax$ groups in its support:}\label{algo:GLqmax}
\begin{algorithmic}[1]
\State{Calculate $\mu_{\text{max}} = \max_v\frac{2}{\sqrt{n}}\Vert K_v^{1/2}(Y-\overline{Y})\Vert$}
\State{Set $\mu_1=\mu_{\text{max}}$ and $\mu_2=\frac{\mu_{\text{max}}}{rat}$} \Comment{"rat" is setted by user.}
\Repeat
 \State{Implement RKHS group lasso algorithm, Algorithm \ref{algo:GL}, with $\mu_i=\frac{\mu_1+\mu_2}{2}$}
  \State{Set $q=\vert \widehat{S}_{\widehat{f}_{\text{Group Lasso}}}\vert$}
   \If{$q>qmax$}
      \State{Set $\mu_1=\mu_1$ and $\mu_2=\mu_i$}
    \Else
      \State{Set $\mu_1=\mu_i$ and $\mu_2=\mu_2$}
    \EndIf 
\Until{$q=qmax$ or $i>$Num} \Comment{"Num" is setted by user.}
\State{Implement RKHS ridge group sparse algorithm, Algorithm \ref{algo:RGS}, with $(\mu=\mu_{q_{max}},\gamma>0)$}
\end{algorithmic}
\end{algorithm}
This algorithm is implemented in the \pkg{RKHSMetaMod} package, as function \fct{RKHSMetMod$\_$qmax} (see Section "Main RKHSMetaMod functions" (supplementary materials) for more details on this function). %This function is described in Section \hyperref[subsec:Main]{Main RKHSMetaMod functions} and illustrated in Example \ref{ex:2}.
%This procedure leads to Algorithm \ref{algo:GLqmax}. This algorithm is implemented in the \pkg{RKHSMetaMod} package, as function \fct{RKHSMetMod$\_$qmax}. This function is described in Section \hyperref[subsec:Main]{Main RKHSMetaMod functions} and illustrated in Example \ref{ex:2}.
\begin{rem}
As both terms in the penalty function of criterion \eref{parametric} enforce sparsity to the solution, the estimator obtained by solving the RKHS ridge group sparse associated with the pair of the tuning parameters $(\mu_{qmax},\gamma>0)$ may contain a smaller number of groups than the solution of the RKHS group lasso optimization problem (i.e. the RKHS ridge group sparse with $(\mu_{qmax},\gamma=0)$). And therefore, the estimated RKHS meta-model contains at most "$qmax$" groups in its support. 
\end{rem}
\vspace{0.3cm}
\section{RKHSMetaMod through examples} \label{sec:examples}
\subsection{Simulation study}\label{subsec:simulstdy}
Let us consider the g-function of Sobol \cite{saltelli2009sensitivity} in the Gaussian regression framework, i.e.
%$$Y=m(X)+\varepsilon,$$
%where the error term $\varepsilon$ is a centered Gaussian random variable with variance $\sigma^2$, and where the function $m$ is the g-function of Sobol defined over $[0,1]^d$ by, 
$Y=m(X)+\varepsilon$. The error term $\varepsilon$ is a centered Gaussian random variable with variance $\sigma^2$, and the function $m$ is the g-function of Sobol defined over $[0,1]^d$ by, 
\begin{align}
\label{gfct}
m(X)=\prod_{a=1}^d\frac{\vert 4X_a-2\vert +c_a}{1+c_a},\:c_a>0 .
\end{align}
The Sobol indices of the g-function can be expressed analytically:
\begin{align*}
\forall v\in\mathcal{P},\:S_v=\frac{1}{D}\prod_{a\in v}D_a,\:D_a=\frac{1}{3(1+c_a)^2},\:D=\prod_{a=1}^d(D_a+1)-1.
\end{align*} 
Set $c_1=0.2$, $c_2=0.6$, $c_3=0.8$ and $(c_a)_{a>3}=100$. With these values of coefficients $c_a$, the variables $X_1, X_2$ and $X_3$ explain $99.98\%$ of the variance of the function $m(X)$ (see Table \ref{trueSI}).

In this Section, three examples are presented. In all examples, the value of Dmax is set as three. Example \ref{ex:1} illustrates the use of the \fct{RKHSMetMod} function by considering three different kernels, "matern", "brownian", and "gaussian" (see Table \ref{kernels}), and three datasets of $n\in\{50,100,200\}$ observations and $d=5$ input variables. %In Example \ref{ex:2}, the function \fct{RKHSMetMod$\_$qmax} is illustrated for the dataset of $n=500$ observations and $d=10$ input variables. 
The larger datasets with $n\in\{1000,2000,5000\}$ observations and $d=10$ input variables are studied in Examples \ref{ex:3} and \ref{ex:4}. 

In each example, two independent datasets are generated: $(X,Y)$ to estimate the meta-models, and $(XT,YT)$ to estimate the prediction errors. The design matrices $X$ and $XT$ are the Latin Hypercube Samples of the input variables that are generated using \fct{maximinLHS} function of the package \pkg{lhs} available at CRAN \cite{lhspackage}:\\%\url{https://CRAN.R-project.org/package=lhs}:\\
\texttt{
R> library(lhs); X <- maximinLHS(n, d); XT <- maximinLHS(n, d)\\
}
The response variables $Y$ and $YT$ are calculated as $Y=m(X)+\varepsilon$ and $YT=m(XT)+\varepsilon_T$, where $\varepsilon$, and $\varepsilon_T$ are centered Gaussian random variables with $\sigma^2=(0.2)^2$:\\%distributed independently according to the centered Gaussian distribution with variance equals to one:
\texttt{
R> a <- c(0.2, 0.6, 0.8, 100, 100, 100, 100, 100, 100, 100)[1:d]\\
R> g=1; for (i in 1:d) g = g*(abs(4*X[,i]-2)+a[i])/(1+a[i])\\
R> sigma <- 0.2 \\
R> epsilon <- rnorm(n, 0, sigma*sigma); Y <- g + epsilon\\
R> gT=1; for (i in 1:d) gT = gT*(abs(4*XT[,i]-2)+a[i])/(1+a[i])\\
R> epsilonT <- rnorm(n, 0, sigma*sigma); YT <- gT + epsilonT\\
}
\begin{exemp}\label{ex:1}
RKHS meta-model estimation using \fct{RKHSMetMod} function:
\end{exemp}
In this example, three datasets of $n$ points \fct{maximinLHS} over $[0,1]^d$ are generated with $n\in\{50,100,200\}$ and $d=5$, and a grid of five values for each of the tuning parameters $\mu$ and $\gamma$ is considered as follows:
\begin{equation*}
\mu_{(1:5)}={\mu_{max}/(\sqrt{n}\times 2^{(2:6)})},\quad \gamma_{(1:5)}=(0.2,0.1,0.01,0.005,0).
\end{equation*}
For each dataset, the experiment is repeated $N_r=50$ times.
At each repetition, the RKHS meta-models associated with the pair of the tuning parameters $(\mu,\gamma)$ are estimated using the \fct{RKHSMetMod} function:\\
\texttt{
R> Dmax <- 3; kernel <- "matern" \# kernel <- "brownian" \# kernel <- "gaussian"\\
R> gamma <- c(0.2, 0.1, 0.01, 0.005, 0); frc <- c(4,8,16,32,64)\\
R> res <- RKHSMetMod(Y, X, kernel, Dmax, gamma, frc, FALSE)\\
}
These meta-models are evaluated using a testing dataset. The prediction errors are computed for them using the \fct{PredErr} function. The RKHS meta-model with minimum prediction error is chosen to be the \textit{best} estimator for the model. Finally, the Sobol indices are computed for the \textit{best} RKHS meta-model using the function \fct{SI$\_$emp}:\\
\texttt{
R> Err <- PredErr(X, XT, YT, mu, gamma, res, kernel, Dmax)\\
R> SI <- SI\_emp(res, Err)\\
}
The vector "mu" is the values of the tuning parameter $\mu$ that are calculated throughout the function \fct{RKHSMetMod} (see argument "frc" in Table \ref{metmod}). It could be recovered from the output of the \fct{RKHSMetMod} function as follows:\\ 
\texttt{
R> mu <- vector()\\
R> l <- length(gamma); for(i in 1:length(frc)){mu[i] <- res[[(i-1)*l+1]]\$mu}\\
}
The performances of this method for estimating a meta-model are evaluated by considering a third dataset $(m(X^{third}_i),X_i^{third})$, $i=1,...,N$, with $N=1000$. The global prediction error is calculated as follows:

Let $\widehat{f}_r(.)$ be the \textit{best} RKHS meta-model obtained in the repetition $r$, $r=1,...,N_r$, then
\begin{align*}
GPE=\frac{1}{N_{r}}\sum_{r=1}^{N_{r}}\frac{1}{N}\sum_{i=1}^N(\widehat{f}_r(X^{third}_i)-m(X^{third}_i))^2.
\end{align*}
The values of $GPE$ obtained for different kernels and values of $n$ are given in Table \ref{tab:errorex1}.
\begin{table}[h!]%[htbp]%[h!]
\centering
\begin{tabular}{l|lll} 
$n$         & $50$     & $100$   & $200$ \\ \hline
$GPE_m$     & $0.13$   & $0.07$  & $0.03$ \\
$GPE_b$     & $0.14$   & $0.10$  & $0.05$ \\
$GPE_g$     & $0.15$   & $0.10$  & $0.07$ \\
\end{tabular}
\caption{Example \ref{ex:1}: The columns of the table correspond to the different datasets with $n\in\{50,100,200\}$ and $d=5$. Each line of the table, from up to down, gives the value of GPE obtained for each dataset associated with the "matern", "brownian" and "gaussian" kernels, respectively.}
\label{tab:errorex1}
\end{table} 
As expected the value of $GPE$ decreases as $n$ increases. The lowest values of $GPE$ are obtained when using the "matern" kernel.

In order to sum up the behaviour of the procedure for estimating the Sobol indices, %the mean square error (MSE) is estimated as follows: 
%Let
%\begin{align*}
%b^2_v=(\widehat{S}_{v,.}-S_v)^2\quad\text{and}\quad w^2_v=\frac{1}{N_r}\sum_{r=1}^{N_r}(\widehat{S}_{v,r}-\widehat{S}_{v,.})^2,
%\end{align*}
%where for each group $v$, $S_v$ denotes the true values of the Sobol indices, and for $\widehat{S}_{v,r}$ being the empirical Sobol indices of the \textit{best} RKHS meta-model in repetition $r$, $\widehat{S}_{v,.}$ denotes the mean of the empirical Sobol indices of the \textit{best} RKHS meta-models through all repetitions: 
%\begin{align*}
%\widehat{S}_{v,.}=\frac{1}{N_r}\sum_{r=1}^{N_r}\widehat{S}_{v,r}.
%\end{align*}
%Then, 
%\begin{align*}
%MSE=\sum_v(b^2_v+w^2_v).
%\end{align*}
we consider the mean square error (MSE) criterion obtained by $\sum_v(\sum_{r=1}^{N_r}(\widehat{S}_{v,r}-S_{v})^2/{N_r})$,
%\begin{align*}
%MSE=\sum_v\frac{1}{N_r}\sum_{r=1}^{N_r}(\widehat{S}_{v,r}-S_{v})^2,
%\end{align*}
where for each group $v$, $S_v$ denotes the true values of the Sobol indices, and $\widehat{S}_{v,r}$ is the empirical Sobol indices of the \textit{best} RKHS meta-model in repetition $r$.

The obtained values of MSE for different kernels and values of $n$ are given in Table \ref{tab:bvarex1}. 
\begin{table}[h!]%[htbp]%[h!]
\centering
\begin{tabular}{l|lll} 
$n$         & $50$           & $100$         & $200$ \\ \hline
$MSE_m$     & $75.12$   & $46.72$  & $28.22$ \\
$MSE_b$     & $110.71$  & $84.99$  & $41.06$ \\
$MSE_g$     & $78.22$   & $94.67$  & $67.02$ \\
\end{tabular}
\caption{Example \ref{ex:1}: The columns of the table correspond to the different datasets with $n\in\{50,100,200\}$ and $d=5$. Each line of the table, from up to down, gives the value of MSE  obtained for each dataset associated with the "matern", "brownian" and "gaussian" kernels, respectively.}
\label{tab:bvarex1}
\end{table} 
As expected, the values of MSE are smaller for larger values of $n$. The smallest values are obtained when using "matern" kernel.

The means of the empirical Sobol indices of the \textit{best} RKHS meta-models through all repetitions for $n=200$ and "matern" kernel are displayed in Table \ref{trueSI}. 
\begin{table}[h!]%[htbp]%[h!]
\centering
\begin{tabular}{l|lllllll|l} 
 $v$ & $\{1\}$ & $\{2\}$ & $\{3\}$ & $\{1,2\}$ & $\{1,3\}$ & $\{2,3\}$ & $\{1,2,3\}$ & sum\\ \hline
 $S_v$ & 43.30 & 24.30 & 19.20 & 5.63 & 4.45 & 2.50 & 0.57 & 99.98\\ \hline
 $\widehat{S}_{v,.}$  & 46.10 & 26.33 & 20.62 & 2.99 & 2.22 & 1.13 & 0.0 & 99.39\\                                                                 
\end{tabular}
\caption{Example \ref{ex:1}: The first line of the table gives the true values of the Sobol indices
$\times100$. The second line gives the mean of the estimated empirical Sobol indices ($\widehat{S}_{v,.}=\sum_{r=1}^{N_r}\widehat{S}_{v,r}/N_r$) $\times100$ greater than $10^{-2}$ calculated over fifty simulations for $n=200$ and "matern" kernel. The sum of the Sobol indices is displayed in the last column.}
\label{trueSI}
\end{table}  
It appears that the estimated Sobol indices are close to the true ones, nevertheless, they are overestimated for the main effects, i.e. groups $v\in\{\{1\},\{2\},\{3\}\}$, and underestimated for the interactions of order two and three, i.e. groups $v\in\{\{1,2\},\{1,3\},\{2,3\},\{1,2,3\}\}$.

Note that the strategy of choosing the tuning parameters is based on the minimization of the prediction error of the estimated meta-model, which may not minimize the error of estimating the Sobol indices.

Taking into account the results obtained for this Example \ref{ex:1}, the calculations in the rest of the examples is done using only the "matern" kernel.

\begin{exemp}\label{ex:3} A time-efficient strategy to obtain the "optimal" tuning parameters when dealing with large data sets:
%A time saving trick to obtain the "optimal" tuning parameters when dealing with larger datasets:
\end{exemp}
A dataset of $n$ points \fct{maximinLHS} over $[0,1]^d$ with $n=1000$ and $d=10$ is generated. %Firstly, the eigenvalues and eigenvectors of the positive definite matrices $K_v$, and the value of $\mu_{max}$ is computed using functions \fct{calc$\_$Kv} and \fct{mu$\_$max} respectively:
First, we use functions \fct{calc$\_$Kv} and \fct{mu$\_$max} to compute the eigenvalues and eigenvectors of the positive definite matrices $K_v$, and the value of $\mu_{max}$, respectively:\\
\texttt{
R> kernel <- "matern"; Dmax <- 3\\
R> Kv <- calc\_Kv(X, kernel, Dmax, TRUE, TRUE)\\
R> mumax <- mu\_max(Y, Kv\$kv)\\
}
Then, we consider the two following steps:
\begin{itemize}
\item[1.] Set $\gamma=0$ and, $\mu_{(1:9)}=\mu_{max}/(\sqrt{n}\times 2^{(2:10)}).$ Calculate the RKHS meta-models associated with the values of $\mu_g=\mu\times\sqrt{n}$ by using the function \fct{RKHSgrplasso}. Gather the obtained RKHS meta-models in a list, "res$\_$g" (while this job could be done with the function \fct{RKHSMetMod} by setting $\gamma=0$, in this example, we use the function \fct{RKHSgrplasso} in order to avoid the re-calculation of $K_v$'s at the next step). Thereafter, for each estimator in res$\_$g, the prediction error is calculated by considering a new dataset and using the function \fct{PredErr}. The value of $\mu$ with the smallest error of prediction in this step is denoted by $\mu_{i}$. Let us implement this step:\\
For a grid of values of $\mu_g$, a sequence of the RKHS meta-models are calculated and gathered in the "res$\_$g" list:\\
\texttt{
R> frc <- c(4,8,16,32,64,128,256,512,1024)\\
R> mu$\_$g <- mumax/frc\\
R> res$\_$g <- list();resg <- list()\\
R> for(i in 1:length(mu$\_$g)){\\
R> resg[[i]] <- RKHSgrplasso(Y,Kv, mu$\_$g[i],1000,FALSE)\\
R> res$\_$g[[i]] <- list("mu$\_$g"=mu$\_$g,"gamma"=0,"MetaModel"=resg[[i]])\\
R> }\\
}
Output res$\_$g contains nine RKHS meta-models and they are evaluated using a testing dataset:\\
\texttt{
R> gamma <- c(0); Err$\_$g <- PredErr(X, XT, YT, mu$\_$g, gamma, res$\_$g, kernel, Dmax)\\
}
The prediction errors of the RKHS meta-models obtained in this step are displayed in Table \ref{tab:errorgex3}.
\begin{table}[h!]%[htbp]%[h!]
\centering
\begin{tabular}{l|lllllllll} 
$\mu_g$    & $1.304$ & $0.652$ & $0.326$ & $0.163$ & $0.081$ & $0.041$ & $0.020$ & $0.010$ & $0.005$\\ \hline
$\gamma=0$ & $0.197$ & $0.156$ & $0.145$ & $0.097$ & $0.063$ & $0.055$&$0.056$ &$0.063$ & $0.073$
\end{tabular}
\caption{Example \ref{ex:3}: Obtained prediction errors in step 1.}
\label{tab:errorgex3}
\end{table} 
It appears that the minimum prediction error corresponds to the solution of the RKHS group lasso algorithm with $\mu_g=0.041$, so $\mu_i=0.041/\sqrt{n}$.
\item[2.] Choose a smaller grid of values of $\mu$, $(\mu_{(i-1)},\mu_i,\mu_{(i+1)})$, and set a grid of values of $\gamma>0$. Calculate the RKHS meta-models associated with each pair of the tuning parameters $(\mu,\gamma)$ by the function \fct{pen$\_$MetMod}. Calculate the prediction errors for the new sequence of the RKHS meta-models using the function \fct{PredErr}. Compute the empirical Sobol indices for the \textit{best} estimator. 
Let us go back to the implementation of the example and apply this step $2$:\\
The grid of the values of $\mu$ in this step is $(0.081,0.041,0.020)/\sqrt{n}.$
The RKHS meta-models associated with this grid of the values of $\mu$ are gathered in a new list "resgnew". We set $\gamma_{(1:4)}=(0.2, 0.1, 0.01, 0.005)$, and we calculate the RKHS meta-models for this new grid of the values of $(\mu,\gamma)$ using \fct{pen$\_$MetMod} function:\\
\texttt{
R> gamma <- c(0.2, 0.1, 0.01, 0.005); mu <- c(mu$\_$g[5], mu$\_$g[6], mu$\_$g[7])/sqrt(n)\\
R> resgnew <- list()\\ 
R> resgnew[[1]] <- resg[[5]]; resgnew[[2]] <- resg[[6]]; resgnew[[3]] <- resg[[7]]\\  
R> res <- pen$\_$MetMod(Y, Kv, gamma, mu, resgnew, 0, 0)\\
}
The output "res" is a list of twelve RKHS meta-models. These meta-models are evaluated using a new dataset, and their prediction errors are displayed in Table \ref{tab:errorex3}.
\begin{table}[h!]%[htbp]%[h!]
\centering
\begin{tabular}{l|lll} 
$\mu$         & $0.081/\sqrt{n}$& $0.041/\sqrt{n}$& $0.020/\sqrt{n}$ \\ \hline
$\gamma=0.2$  & $0.153$ & $0.131$ & $0.119$ \\
$\gamma=0.1$  & $0.098$ & $0.079$ & $0.072$ \\
$\gamma=0.01$ & $0.065$ & $0.054$ & $0.053$ \\
$\gamma=0.005$& $0.064$ & $0.054$ & $0.054$ \\
\end{tabular}
\caption{Example \ref{ex:3}: Obtained prediction errors in step 2.}
\label{tab:errorex3}
\end{table} 

The minimum prediction error is associated with the pair $(0.020/\sqrt{n},0.01)$, and the \textit{best} RKHS meta-model is then $\widehat{f}_{(0.020/\sqrt{n},0.01)}$. 

The performance of this procedure for estimating the Sobol indices is evaluated using the relative error (RE) defined as follows:\\
For each $v$, let $S_v$ be the true value of the Sobol indices displayed in Table \ref{trueSI} and $\widehat{S}_v$ be the estimated empirical Sobol indices. Then
\begin{equation}
\label{relerr}
RE=\sum_v{\vert\widehat{S}_v-S_v\vert/S_v}.
\end{equation}
In Table \ref{tab:SIex3} the estimated empirical Sobol indices, their sum, and the value of RE are displayed.  
\begin{table}[h!]%[htbp]%[h!]
\centering
\begin{tabular}{l|lllllll|l|l} 
$v$    & $\{1\}$ & $\{2\}$ & $\{3\}$ & $\{1,2\}$ & $\{1,3\}$ & $\{2,3\}$ & $\{1,2,3\}$ & sum &RE \\ \hline
$\widehat{S}_v$   & $42.91$  & $25.50$  & $20.81$  & $4.40$ & $3.84$ & $2.13$ & $0.00$ & $99.60$ & $1.64$
\end{tabular}
\caption{Example \ref{ex:3}: The estimated empirical Sobol indices $\times100$ greater than $10^{-2}$. The last two columns show $\sum_v\widehat{S}_v$ and RE, respectively.}
\label{tab:SIex3}
\end{table} 
\end{itemize}
The obtained RE for each group $v$ is smaller than $1.64\%$, therefore, the estimated Sobol indices in this example are very close to the true values of the Sobol indices displayed in the first row of Table \ref{trueSI}. %In this example the significant values of the Sobol indices for interactions of order two are obtained.

\begin{figure}[h!]%[htbp]
  \centering
  \includegraphics[width=10cm]{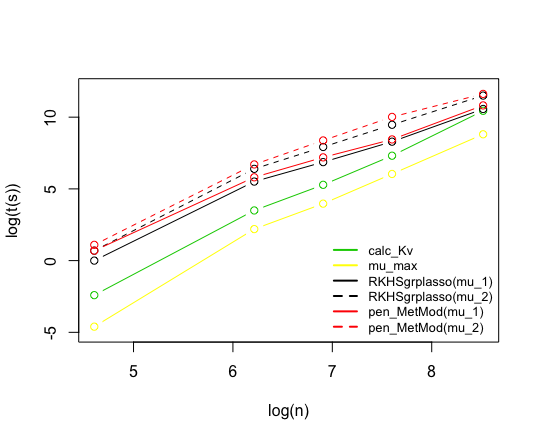}
  \caption{Example \ref{ex:4}: Timing plot for $d=10$, $n\in\{100,300,500,1000,2000,5000\}$, and different functions of the \pkg{RKHSMetaMod} package. The logarithm of the execution time (in seconds) for the functions \fct{RKHSgrplasso} and \fct{pen$\_$MetMod} is displayed for two pairs of values of tuning parameters $(\mu_1=\mu_{max}/(\sqrt{n}\times 2^7),\gamma=0.01)$ in solid lines, and $(\mu_2=\mu_{max}/(\sqrt{n}\times 2^8),\gamma=0.01)$ in dashed lines.}
  \label{timingplot}
  %\end{center}
\end{figure}

\begin{exemp}\label{ex:4}
Dealing with larger datasets:
\end{exemp} 
Two datasets of $n$ points \fct{maximinLHS} over $[0,1]^d$ with $n\in\{2000,5000\}$ and $d=10$ are generated.
In order to obtain one RKHS meta-model associated with one pair of the tuning parameters $(\mu,\gamma)$, the number of coefficients to be estimated is equal to $n\times$vMax$=n\times 175$. 
Table \ref{timing} gives the execution time for different functions used throughout the Examples \ref{ex:1}-\ref{ex:4}. In all examples we used a cluster of computers with: 2 Intel Xeon E5-2690 processors (2.90GHz) and 96Gb Ram (6x16Gb of memory 1600MHz). 
\begin{table}[h!]%[htbp]%[h!]
\centering
\begin{tabular}{l|llllll}
 $(n,d)$ & {\small\fct{calc$\_$Kv}} & {\small\fct{mu$\_$max}} & {\small\fct{RKHSgrplasso}} & {\small\fct{pen$\_$MetMod}} & $\vert S_{\widehat{f}}\vert$ & {\small sum} \\ \hline 
\multirow{2}{*}{(100,5)} & \multirow{2}{*}{0.09s} & \multirow{2}{*}{0.01s} & 1s & 2s & 18 & $\sim$ 3s\\ 
                                       &  &  & 2s & 3s & 19 & $\sim$ 5s \\ \hline
\multirow{2}{*}{(500,10)} & \multirow{2}{*}{33s} & \multirow{2}{*}{9s} & 247s & 333s & 39 & $\sim$ 10min\\ 
                                        & &  & 599s & 816s & 64 & $\sim$ 24min\\ \hline
\multirow{2}{*}{(1000,10)} & \multirow{2}{*}{197s} & \multirow{2}{*}{53s} & 959s &  1336s & 24 & $\sim$ 42min\\
                                    &   &  & 2757s & 4345s &  69 & $\sim$ 2h \\ \hline
\multirow{2}{*}{(2000,10)} & \multirow{2}{*}{1498s} & \multirow{2}{*}{420s} & 3984s & 4664s & 12 & $\sim$ 2h:56min\\ 
                                     &  &  & 12951s & 22385s & 30 &  $\sim$ 10h:20min\\ \hline
\multirow{2}{*}{(5000,10)} & \multirow{2}{*}{34282s} & \multirow{2}{*}{6684s} & 38957s & 49987s & 11 & $\sim$ 36h:05min\\ 
                                     &  &  & 99221s & 111376s & 15 & $\sim$ 69h:52min\\ 
\end{tabular}
\caption{Example \ref{ex:4}: The kernel used is "matern". The execution time for the functions \fct{RKHSgrplasso} and \fct{pen$\_$MetMod} is displayed  in each row for two pairs of values of tuning parameters $(\mu_1=\mu_{max}/(\sqrt{n}\times 2^7),\gamma=0.01)$ on up, and $(\mu_2=\mu_{max}/(\sqrt{n}\times 2^8),\gamma=0.01)$ on below. In the column $\vert S_{\widehat{f}}\vert$, the  number of the active groups associated with each estimated RKHS meta-model is displayed.}\label{timing}
\end{table}
As we can see, the execution time increases fast as $n$ increases. In Figure \ref{timingplot} the plot of the logarithm of the time (in seconds) versus the logarithm of $n$ is displayed for the functions \fct{calc$\_$Kv}, \fct{mu$\_$max}, \fct{RKHSgrplasso} and \fct{pen$\_$MetMod}.
%\begin{figure}[h!]
%\begin{center}
It appears that, the algorithms of these functions are of polynomial time $O(n^\alpha)$ with $\alpha\backsimeq3$ for the functions \fct{calc$\_$Kv} and \fct{mu$\_$max}, and $\alpha\backsimeq2$ for the functions \fct{RKHSgrplasso} and \fct{pen$\_$MetMod}.

Taking into account the results obtained for the prediction error and the values of $(\widehat{\mu},\widehat{\gamma})$ in Example \ref{ex:3}, in this example, only two values of the tuning parameter $\mu_{(1:12)}=\mu_{max}/(\sqrt{n}\times2^{(7:8)})$, and one value of the tuning parameter $\gamma=0.01$ are considered.
%\begin{align*}
%\mu=(\frac{\mu_{max}}{(\sqrt{n}\times 2^{7})}, \frac{\mu_{max}}{(\sqrt{n}\times 2^{8})})\mbox{ and }\gamma=0.01.
%\end{align*}
The RKHS meta-models associated with the pair of values $(\mu_i,\gamma)$, $i=1,2$ are estimated using the \fct{RKHSMetMod} function:\\
\texttt{
R> kernel <- "matern"; Dmax <- 3\\
R> gamma <- c(0.01); frc <- c(128,256)\\
R> res <- RKHSMetMod(Y, X, kernel, Dmax, gamma, frc, FALSE)\\
}
The prediction error and the empirical Sobol indices are then calculated for the obtained meta-models using the functions \fct{PredErr} and \fct{SI$\_$emp}:\\
\texttt{
R> mu <- vector(); mu[1] <- res[[1]]\$mu; mu[2] <- res[[2]]\$mu\\
R> Err <- PredErr(X, XT, YT, mu, gamma, res, kernel, Dmax)\\
R> SI <- SI$\_$emp(res, NULL)\\
}
The result of the prediction errors associated with the obtained estimators for two different values of $n$ are displayed in Table \ref{tab:errorex4}. 
\begin{table}[h!]%[htbp]%[h!]
\centering
\begin{tabular}{l|ll} 
 $n$ & $(\mu_{max}/(\sqrt{n}\times 2^{7}),\gamma)$ & $(\mu_{max}/(\sqrt{n}\times 2^{8},\gamma)$ \\ \hline
$2000$  & $0.052$ & $0.049$  \\ \hline
$5000$  & $0.049$ & $0.047$  \\
\end{tabular}
\caption{Example \ref{ex:4}: Obtained prediction errors.}
\label{tab:errorex4}
\end{table} 
For $n$ equal to $5000$, we obtained smaller values of the prediction error, so as expected, the prediction quality improves by increasing the number of the observations $n$. 
Table \ref{tab:SIex4} gives the estimated empirical Sobol indices as well as their sum and the values of RE (see Equation \eref{relerr}).
\begin{table}[h!]%[htbp]%[h!]
\centering
\begin{tabular}{l|l|lllllll|l|l} 
$n$ & $v$    & $\{1\}$ & $\{2\}$ & $\{3\}$ & $\{1,2\}$ & $\{1,3\}$ & $\{2,3\}$ & $\{1,2,3\}$ & sum & RE \\ \hline
\multirow{2}{2em}{$2000$}  & $\widehat{S}_{v;(\mu_1,\gamma)}$  & $45.54$  & $24.78$  & $21.01$  & $3.96$     & $3.03$     & $1.65$     & $0.00$       & $99.97$ & $2.12$\\

& $\widehat{S}_{v;(\mu_2,\gamma)}$   & $45.38$  & $25.07$  & $19.69$  & $4.36$     & $3.66$     & $1.79$     & $0.00$       & $ 99.95$ & $1.79$ \\ \hline

\multirow{2}{2em}{$5000$}  & $\widehat{S}_{v;(\mu_1,\gamma)}$  & $44.77$  & $25.39$  & $20.05$  & $4.49$     & $3.38$     & $1.90$     & $0.00$       & $99.98$ & $1.81$\\

& $\widehat{S}_{v;(\mu_2,\gamma)}$   & $43.78$  & $24.99$  & $19.56$  & $5.43$     & $3.90$     & $2.32$     & $0.00$       & $99.98$ & $1.29$
\end{tabular}
\caption{Example \ref{ex:4}: The estimated empirical Sobol indices $\times100$ greater than $10^{-2}$ associated with each estimated RKHS meta-model is printed. The last two columns show $\sum_v\widehat{S}_v$ and RE, respectively. We have $\mu_1=\mu_{max}/(\sqrt{n}\times 2^{7})$, $\mu_2=\mu_{max}/(\sqrt{n}\times 2^{8})$ and $\gamma=0.01$.}
\label{tab:SIex4}
\end{table} 
Comparing the values of RE, we can see that the empirical Sobol indices are better estimated for n equal to $5000$, so as expected, the estimation of the Sobol indices is better for larger values of $n$.

\begin{figure}[h!]%[htbp]
  \centering
  \includegraphics[height=7cm,width=10cm]{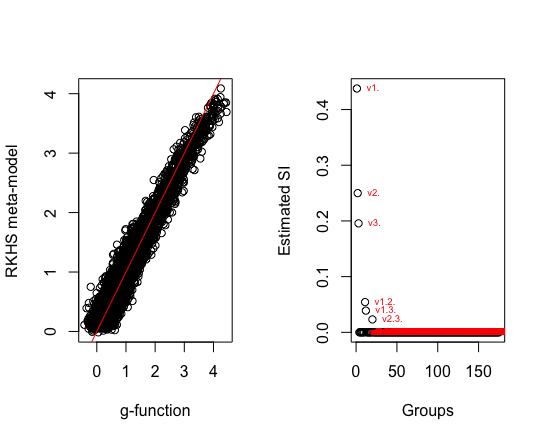}
  \caption{Example \ref{ex:4}: On the left, the RKHS meta-model versus the g-function is plotted. On the right, the empirical Sobol indices in the $y$ axis and vMax$=175$ groups in the $x$ axis are displayed.}
  \label{SI}
  %\end{center}
\end{figure}

In Figure \ref{SI} the result of the estimation quality and the Sobol indices for the dataset with $n$ equal to $5000$, $d$ equal to $10$, and $(\mu_2,\gamma)$ are displayed.
%\begin{figure}[h!]
%\begin{center}
The line $y = x$ in red crosses the cloud of points as long as the values of the g-function are smaller than three. 
When the values of the g-function are greater than three, the estimator $\widehat{f}$ tends to underestimate the g-function.
%% -- Summary/conclusions/discussion -------------------------------------------
%$
\subsection{Comparison examples}\label{subsec:comprexpl}
This section includes two examples. In the first example we reproduce an example from paper \cite{Gu_2019} and compare the prediction quality of the RKHS meta-model with the GP (Kriging) based meta-models from the \pkg{RobusGaSP} \cite{Gu_2019} and \pkg{DiceKriging} \cite{JSSv051i01} packages. 

The objective is to evaluate the quality of the
RKHS estimated meta-model and to compare it with methods recently
proposed for approximating complex models. In the first example we
consider one-dimensional model and focuse on the comparison between the
true  model and the estimated meta-model. 
In the second example we reproduce an example from paper \cite{JSSv051i01} which allows us to compare the prediction quality of the RKHS meta-model with the Kriging based meta-model from \pkg{DiceKriging} package, as well as the estimation quality of the Sobol indices in our package with the well-known package \pkg{sensitivity}.  

\begin{figure}[h!]%[htbp]
  \centering
  \includegraphics[width=10cm]{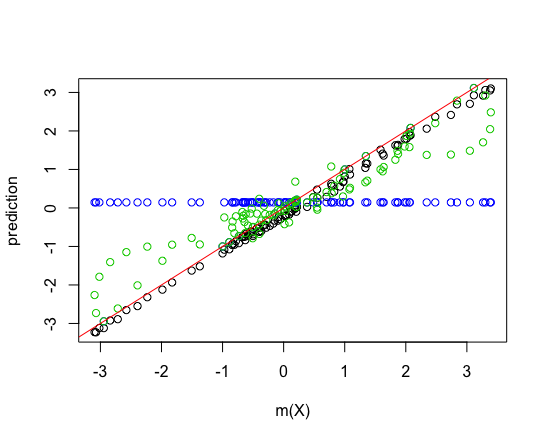}
  \caption{Example \ref{ex:comparisonPrediction}: Prediction of the modified sine wave function with $100$ test points, equally spaced in $[0,1]$. The x-axis is the real output and the y-axis is the prediction. The black circles are the prediction from \pkg{RKHSMetMod}, the green circles are the predictive mean from \pkg{RobustGaSP}, and the blue circles are the predictive mean from \pkg{DiceKriging}.}
  \label{Rplot01ex6}
  %\end{center}
\end{figure}

\begin{figure}[h!]%[htbp]
  \centering
  \includegraphics[width=10cm]{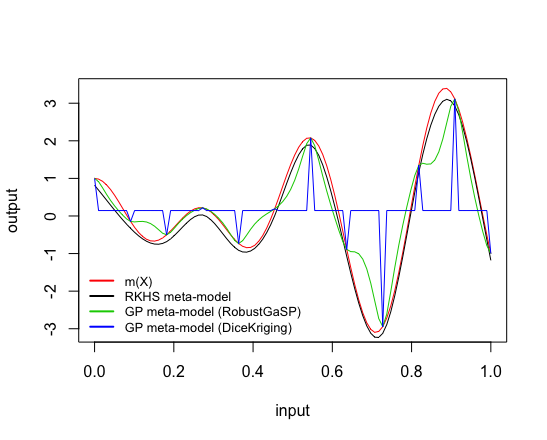}
  \caption{Example \ref{ex:comparisonPrediction}: The red curve is the graph of the modified sine wave function with $100$ test points, equally spaced in $[0,1]$. The black curve is the prediction produced by the
\pkg{RKHSMetaMod} package. The blue curve is the predictive mean by the \pkg{DiceKriging} package, and the green curve is the predictive mean produced by the \pkg{RobustGaSP} package.}
  \label{Rplotex6}
  %\end{center}
\end{figure}

For the sake of comparison between the three methods, the meta-models  are calculated using
the same experimental design and outputs and the same kernel function
available in
 three packages is used. However, in packages \pkg{RobustGaSP} and \pkg{DiceKriging} the range parameter $r$ (see Table \ref{kernels}) in the kernel function is estimated by marginal posterior modes with the robust parameterization and by MLE with upper and lower bounds, respectively, while it is assumed to be fixed and set as $\sqrt{3}/2$ in the \pkg{RKHSMetaMod} package.  
\begin{exemp}\label{ex:comparisonPrediction} "The modified sine wave function" \cite{Gu_2019}.
\end{exemp}
We consider the $1$-dimensional modified sine wave function defined over $[0,1]$ by:
\begin{align*}
m(X)=3\mbox{sin}(5\pi X)+\mbox{cos}(7\pi X).
\end{align*}
The same experimental design as described in \cite{Gu_2019} is considered: the design matrix $X$ is a sequence of $12$ equally spaced points on $[0,1]$, and the response variable $Y$ is calculated as $Y = m(X)$:\\
\texttt{
R> X <- as.matrix(seq(0,1,1/11)); Y <- sinewave(X)\\
}
where \fct{sinewave} function is defined in \cite{Gu_2019}. 
We build the GP based meta-models by the \pkg{RobustGaSP} and the \pkg{DiceKriging} packages using the constant mean function and kernel Matérn $3/2$:\\
\texttt{
R> library(RobustGaSP)\\
R> res.rgasp <- rgasp(design=X, response=Y, kernel$\_$type="matern$\_$3$\_$2")\\
R> library(DiceKriging)\\
R> res.km <- km(design=X, response=Y, covtype="matern3$\_$2")\\
} 
As $d=1$, we have $Dmax=1$. We consider the grid of values of $\mu_{(1:9)}=\mu_{max}/(\sqrt{n}\times 2^{(2:10)})$ and $\gamma_{(1:5)}=(0.2,0.1,0.01,0.005,0)$.
%\begin{equation*}
%\mu_{(1:9)}=\frac{\mu_{max}}{(\sqrt{n}\times 2^{(2:10)})},\quad \gamma_{(1:5)}=(0.2,0.1,0.01,0.005,0).
%\end{equation*}
%Note that the large grid of values of $\mu$ does not cause a problem in this example since we deal with a small dataset. In order to choose the appropriate grid of values of $\mu$ when dealing with large datasets one may consider the same strategy as described in Example \ref{ex:3} of the previous Section. 
The RKHS meta-models associated with the pair of values $(\mu_i,\gamma_j)$, $i = 1,\cdots,9$, $j=1,\cdots,5$ are estimated using the \fct{RKHSMetMod} function:\\
\texttt{
R> kernel <- "matern"; Dmax <- 1\\ 
R> gamma <- c(0.2, 0.1, 0.01, 0.005,0)\\
R> frc <- c(4,8,16,32,64,128,256,512,1024)\\
R> res <- RKHSMetMod(Y, X, kernel, Dmax, gamma, frc, FALSE)\\
}
Given a testing dataset $(XT,YT)$, the prediction errors associated with the obtained RKHS meta-models are calculated using the \fct{PredErr} function, and the \textit{best} RKHS meta-model is chosen to be the estimator of the model $m(X)$:\\
\texttt{
R> XT <- as.matrix(seq(0,1,1/11)); YT <- sinewave(XT)\\
R> Err <- PredErr(X, XT, YT, mu, gamma, res, kernel, Dmax)\\
}
To compare these three estimators in terms of the prediction quality, we perform prediction on $100$ test points, equally spaced in $[0,1]$:\\
\texttt{
R> predict$\_$X <- as.matrix(seq(0,1,1/99))\\
R> \#prediction with the GP based meta-models:\\
R> rgasp.predict <- predict(res.rgasp, predict$\_$X)\\
R> km.predict <- predict(res.km, predict$\_$X, type='UK')\\
R> \#prediction with the best RKHS meta-model:\\
R> res.predict <- prediction(X, predict$\_$X, kernel, Dmax, res, Err)\\ 
}
The prediction results are plotted in Figure \ref{Rplot01ex6}. The black circles that correspond to the prediction from the \pkg{RKHSMetMod} package are closer to the real output than the green and the blue circles corresponding to the predictive means from the \pkg{RobustGaSP} and \pkg{DiceKriging} packages.
The meta-model results are plotted in Figure \ref{Rplotex6}. The prediction from the \pkg{RKHSMetaMod} package plotted as the black curve is much more accurate as an estimate of the true function (plotted in red) than the predictive mean from the \pkg{RobustGaSP} and \pkg{DiceKriging} packages plotted as the blue and green curves, respectively. As already noted by \cite{Gu_2019}, for that sine wave
example, the meta-model from the DiceKriging package "degenerates to the
fitted mean with spikes  at the design points".

\begin{exemp}\label{ex:comparison} "A standard SA 8-dimensional example" \cite{JSSv051i01}.
\end{exemp}
We consider the $8$-dimensional g-function of Sobol implemented in the package \pkg{sensitivity}: the function $m(X)$ as defined in Equation \eref{gfct} with coefficients $c_1=0,\:c_2=1,\:c_3=4.5,\:c_4=9,\:(c_a)_{a=5,6,7,8}=99$.
%$$m(X)=\prod_{a=1}^d\frac{\vert 4x_a-2\vert +c_a}{1+c_a},\: X\in[0,1]^8.$$
With these values of coefficients $c_a$, the variables $X_1$, $X_2$, $X_3$ and $X_4$ explain $99.96\%$ of the variance of the function $m(X)$ (see Table \ref{trueSIcomparison}).
  
We consider the same experimental design as described in \cite{JSSv051i01}: the design matrices $X$ and $XT$ are $80$-point optimal Latin Hypercube Samples of the input variables generated by the \fct{optimumLHS} function of package \pkg{lhs}, and the response variables $Y$ and $YT$ are calculated as $Y = m(X)$, and $YT = m(XT)$ using \fct{sobol.fun} function of the package \pkg{sensitivity}:\\
\texttt{
R> n <- 80; d <- 8\\
R> library(lhs); X <- optimumLHS(n, d); XT <- optimumLHS(n, d)\\
R> library(sensitivity); Y <- sobol.fun(X); YT <- sobol.fun(XT)\\
}
Let us first consider the RKHS meta-model method. We set Dmax$=3$, and we consider the grid of values of $\mu_{(1:9)}=\mu_{max}/(\sqrt{n}\times 2^{(2:10)})$, and $\gamma_{(1:5)}=(0.2,0.1,0.01,0.005,0)$.  
The RKHS meta-models associated with the pair of values $(\mu_i,\gamma_j)$, $i = 1,\cdots,9$, $j=1,\cdots,5$ are estimated using the \fct{RKHSMetMod} function:\\
\texttt{
R> kernel <- "matern"; Dmax <- 3\\ 
R> gamma <- c(0.2, 0.1, 0.01, 0.005,0)\\
R> frc <- c(4,8,16,32,64,128,256,512,1024)\\ 
R> res <- RKHSMetMod(Y, X, kernel, Dmax, gamma, frc, FALSE)\\
}
Given the testing dataset $(XT,YT)$, the prediction errors associated with the obtained RKHS meta-models are calculated using \fct{PredErr} function, and the \textit{best} RKHS meta-model is chosen to be the estimator of the model $m(X)$. Finally, the Sobol indices are computed for the \textit{best} RKHS meta-model using the function \fct{SI$\_$emp}:\\
\texttt{
R> Err <- PredErr(X, XT, YT, mu, gamma, res, kernel, Dmax)\\
R> SI <- SI$\_$emp(res, Err)\\
}
%In order to reproduce the estimated kriging based meta-model, we use \fct{km} function of package \pkg{DiceKriging}:
Secondly, let us build the GP based meta-model. We use the \fct{km} function of the package \pkg{DiceKriging} with the constant mean function and kernel Matérn $3/2$:\\
\texttt{
R> library(DiceKriging)\\
R> res.km <- km(design = X, response = Y, covtype = "matern3$\_$2")\\
}
The Sobol indices associated with the estimated GP based meta-model are calculated using \fct{fast99} function of the package \pkg{sensitivity}:\\
%kriging.mean <- function(Xnew, m){
%  predict.km(m, Xnew, "UK", se.compute = FALSE, checkNames = FALSE)$mean
%}
\texttt{
R> SI.km <- fast99(model = kriging.mean, factors = d, n = 1000,\\
                      + q = "qunif", q.arg = list(min = 0, max = 1), m = res.km)\\
}
where \fct{kriging.mean} function is defined in \cite{JSSv051i01}. 

\begin{figure}[h!]%[htbp]
  \centering
  \includegraphics[width=10cm]{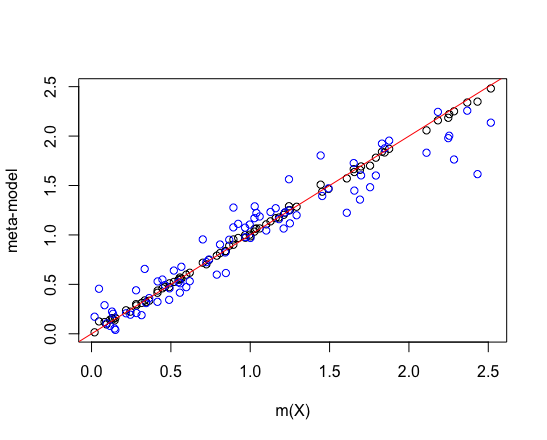}
  \caption{Example \ref{ex:comparison}: The x-axis is the real output and the y-axis is the fitted meta-model. The black circles are the meta-model from \pkg{RKHSMetMod} and the blue circles are the meta-model from \pkg{DiceKriging}.}
  \label{comppred}
  %\end{center}
\end{figure}

%To compare the obtained estimators in terms of the prediction quality, we perform prediction on the testing dataset:
%\begin{example}
%#prediction with the best RKHS meta-model:
%minCritTest <-  which(Err == min(Err, na.rm=TRUE),arr.ind = TRUE)
%nbr <- minCritTest[1]*minCritTest[2]
%res.predict <- prediction(X, XT, kernel, Dmax, res, nbr) 
%#prediction with the GP based meta-model:
%km.predict <- predict(res.km, XT, type='UK')
%\end{example}
The result of the estimation with the best RKHS meta-model and the Kriging based meta-model is drawn in Figure \ref{comppred}. 
%For the kriging based meta-model (the plot on left), the cloud of points are not completely concentrated around the line $y = x$ in red. However, the line $y=x$ crosses the cloud of points as long as the values of the g-function are smaller than two. When the values of the g-function are greater than two, the kriging based estimator tends to under estimate the g-function. For the RKHS meta-model (the plot on right), the cloud of points are completely concentrated around the line $y = x$ in red. In order to be more precise about the prediction quality, one may consider the mean square error of the fitted meta-model computed by $\sum_{i=1}^{80}(m(X_i)-\hat{f}(X_i))^2/80$, where we got $3.96\%$ and $0.07\%$ for the kriging based meta-model and the RKHS meta-model, respectively.  
The black circles that correspond to the best RKHS meta-model are closer to the real output than the blue circles corresponding to the GP based meta-model from the \pkg{DiceKriging} package. Another way to evaluate the prediction quality of the estimated meta-models is to  consider the mean square error of the fitted meta-model computed by $\sum_{i=1}^{80}(m(X_i)-\hat{f}(X_i))^2/80$. We obtained $3.96\%$ and $0.07\%$ for the Kriging based meta-model and the RKHS meta-model, respectively, which confirms the good behavior of the RKHS meta-model. %In order to be more precise about the estimation quality of the meta-models, one may consider the mean square error of the fitted meta-model computed by $\sum_{i=1}^{80}(m(X_i)-\hat{f}(X_i))^2/80$, where we obtained $3.96\%$ and $0.07\%$ for the Kriging based meta-model and the RKHS meta-model, respectively.

The estimated Sobol indices associated with the RKHS meta-model and the Kriging based meta-model are given in Table \ref{trueSIcomparison}.
%\begin{table}[h!]%[htbp]%[h!]
%\centering
%\begin{tabular}{l|lllllllllll|l} 
% $v$ & $\{1\}$ & $\{2\}$ & $\{3\}$ & $\{4\}$ & $\{1,2\}$ & $\{1,3\}$ & $\{1,4\}$ & $\{2,3\}$ & $\{2,4\}$ & $\{1,2,3\}$ &$\{1,2,4\}$ &  sum \\ \hline
% $S_v$ & 71.62 & 17.90 & 2.37 & 0.72 & 5.97 & 0.79 & 0.24 & 0.20 & 0.06 & 0.07 & 0.02& 99.96 \\ 
% $\hat{S}_v$ & 75.78 & 17.42 & 1.71 & 0.47 & 4.00 & 0.05 & 0.07 & 0.28 & 0.09 & 0.00 & 0.00 & 99.87 \\
% $\widehat{S}_{km_v}$ & 71.18 & 15.16 & 1.42 & 0.44 & 0.00 & 0.00 & 0.00 & 0.00 & 0.00 & 0.00 & 0.00 & 88.20 \\
%\end{tabular}
%\caption{Example \ref{ex:comparison}: The true values of the Sobol indices
%$\times100$ greater than $10^{-2}$ are given in the first raw. The estimated Sobol indices associated with the RKHS meta-model ($\hat{S}_v$) and the Kriging based meta-model ($\widehat{S}_{km_v}$) are given in second and third rows, respectively.}
%\label{trueSIcomparison}
%\end{table} 
\begin{table}[h!]
\centering
\small{
{\setlength{\tabcolsep}{7pt}
\begin{tabular}{l|lllllllllll|l} 
 $v$ & $\{1\}$ & $\{2\}$ & $\{3\}$ & $\{4\}$ & $\{1,2\}$ & $\{1,3\}$ & $\{1,4\}$ & $\{2,3\}$ & $\{2,4\}$ & $\{1,2,3\}$ &$\{1,2,4\}$ &  sum \\ \hline
 $S_v$ & 71.62 & 17.90 & 2.37 & 0.72 & 5.97 & 0.79 & 0.24 & 0.20 & 0.06 & 0.07 & 0.02& 99.96 \\ 
 $\hat{S}_v$ & 75.78 & 17.42 & 1.71 & 0.47 & 4.00 & 0.05 & 0.07 & 0.28 & 0.09 & 0.00 & 0.00 & 99.87 \\
 $\widehat{S}_{km_v}$ & 71.18 & 15.16 & 1.42 & 0.44 & 0.00 & 0.00 & 0.00 & 0.00 & 0.00 & 0.00 & 0.00 & 88.20 \\
\end{tabular}}}
\caption{Example \ref{ex:comparison}: The true values of the Sobol indices
$\times100$ greater than $10^{-2}$ are given in the first raw. The estimated Sobol indices associated with the RKHS meta-model ($\hat{S}_v$) and the Kriging based meta-model ($\widehat{S}_{km_v}$) are given in second and third rows, respectively.}
\label{trueSIcomparison}
\end{table} 
As shown, with RKHS meta-model, we obtained non-zero values for the interactions of order two. Concerning the main effects, excepting the first one, the estimated Sobol indices with the RKHS meta-model are closer to the true ones. However, the interactions of order three are ignored by both meta-models. For a general comparison of the estimation quality of the Sobol indices, one may consider the criterion RE defined in Equation \eref{relerr}, which is equal to $7.95$ for the Kriging based meta-model, and $5.59$ for the RKHS meta-model. Comparing the values of RE, we can point out that the Sobol indices are better estimated with the RKHS meta-model in that model.
%Comparing the values of RE, we can point out that the Sobol indices are better estimated with the RKHS meta-model. We got significant values for the interactions of order two with RKHS meta-model and the values of the estimated Sobol indices for main effects $\{3\}$ and $\{4\}$ are closer to the true ones. However, the values of the estimated Sobol indices for main effects $\{1\}$ and $\{2\}$ are closer to the true ones with kriging based meta-model.
\vspace{0.3cm}
\section{Summary and discussion} \label{sec:summary}
In this paper, we proposed an R package, called \pkg{RKHSMetaMod}, that estimates a meta-model of a complex model $m$. This meta-model belongs to a reproducing kernel Hilbert space constructed as a direct sum of Hilbert spaces \cite{DURRANDE201357}. The estimation of the meta-model is carried out via a penalized least-squares minimization allowing both to select and estimate the terms in the Hoeffding decomposition, and therefore, to select the Sobol indices that are non-zero and estimate them \cite{huet:hal-01434895}. This procedure makes it possible to estimate the Sobol indices of high order, a point known to be difficult in practice.
Using the convex optimization tools, \pkg{RKHSMetaMod} package implements two optimization algorithms: the minimization of the RKHS ridge group sparse criterion \eref{parametric} and the RKHS group lasso criterion \eref{Lasso}. 
Both of these algorithms rely on the Gram matrices $K_v$, $v\in\mathcal{P}$ and their positive definiteness. 
Currently, the package considers only uniformly distributed input variables. If one is interested by another distribution of the input variables, it suffices to modify the calculation of the kernels $k_{0a}$, $a=1,...,d$ (see Equation \eref{kernelmodify}) in the function \fct{calc$\_$Kv} of this package (see Remark \ref{ch3:remxun}). 
The available kernels in the \pkg{RKHSMetaMod} package are: Gaussian kernel (with the fixed range parameter $r=1/2$), MatÃ©rn kernel (with the fixed range parameter $r=\sqrt{3}/2$), Brownian kernel, quadratic kernel and linear kernel (see Table \ref{kernels}). With regard to the problem being under study, one may consider other kernels or kernels with different values of the range parameter $r$ and add them easily to the list of the kernels in the \fct{calc$\_$Kv} function.
%a new kernel and add it easily to the list of the kernels in the \fct{calc$\_$Kv} function. Indeed, the choice of different kernels allows us to consider different approximation spaces and choose the one that gives the best result. The range parameters in the Gaussian and MatÃ©rn $3/2$ kernels are assumed to be fixed and equal to a given value (see Table \ref{kernels}). We do not propose any method to estimate these parameters since in our context (high dimensional data) the computation time is very expensive. However, these parameters could be changed and replaced easily by another value in the \fct{calc$\_$Kv} function.    
%With regard to the problem being under study, one may consider another kernel and add it easily to the list of the kernels in the \fct{calc$\_$Kv} function. Indeed, the choice of different kernels allows us to consider different approximation spaces and choose the one that gives the best result.  
For the large values of $n$ and $d$ the calculation and storage of the eigenvalues and the eigenvectors of all the Gram matrices $K_v$, $v\in\mathcal{P}$ require a lot of time and a very large amount of memory. In order to optimize the execution time and also the storage memory, except for a function that is written in R, all of the functions of \pkg{RKHSMetaMod} package are written using the efficient C++ libraries through \pkg{RcppEigen} and \pkg{RcppGSL} packages. These functions are then interfaced with the R environment in order to propose a user friendly package. 
\section{Supplementary materials}
\subsection{Appendix} \label{app:technical}
\begin{mydef}\label{rem:subdiff}
For $F(x)=\Vert Ax\Vert$, where $A$ is a symmetric matrix that does not depend on x, we have: $\partial F(x)=\{{A^2x/\Vert Ax\Vert}\}$ if $x\neq 0$, and $\partial F(x)=\{w\in\mathbb{R}^n,\quad\Vert A^{-1}w\Vert\leq 1\}$ if $x=0$.
%\begin{align*}
%&\partial F(x)=\{\frac{A^2x}{\Vert Ax\Vert}\}\quad &\text{if}\quad x\neq 0,\\
%&\partial F(x)=\{w\in\mathbb{R}^n,\quad\Vert A^{-1}w\Vert\leq 1\}\quad &\text{if}\quad x=0.
%\end{align*}
\end{mydef}
\begin{mydef}\label{fooc}
Let $F:\mathbb{R}^n\rightarrow \mathbb{R}$ be a convex function. We have the following first order optimality condition:
$$\widehat{x}\in \text{argmin}_{x\in\mathbb{R}^n}F(x)\Leftrightarrow 0\in\partial F(\widehat{x}).$$
This results from the fact that $F(y)\geq F(\widehat{x})+ <0,y-\widehat{x}> $ for all $y\in\mathbb{R}^n$ in both cases \cite{giraud2014introduction}.
\end{mydef}
\subsubsection{RKHS construction}\label{subsec:Data}
We begin this Section with a brief introduction to the RKHS.
Let $\mathcal{H}$ be a Hilbert space of real valued functions on a set $\mathcal{X}$. The space $\mathcal{H}$ is a RKHS if for all $X\in\mathcal{X}$ the evaluation functionals $L_X:\mathcal{H}(f)\rightarrow \mathbb{R}(f(X))$  
are continuous. The Riesz representation Theorem ensures the existence of an unique element $k_X(.)$ in $\mathcal{H}$ verifying the property that $
\forall X\in\mathcal{X},\:\forall f\in\mathcal{H},\: f(X)=L_X(f)=\langle f,k_X\rangle_{\mathcal{H}},$
where $\langle.,.\rangle_{\mathcal{H}}$ denotes the inner product in $\mathcal{H}$.
It follows that for all $X$ ,$X'$ in $\mathcal{X}$, and $k_X(.)$, $k_{X'}(.)$ in $\mathcal{H}$, we have
$k_X(X')=L_{X'}(k_X)=\langle k_X,k_{X'}\rangle_{\mathcal{H}}.$
This allows to define the reproducing kernel of $\mathcal{H}$ as $
k:\mathcal{X}\times\mathcal{X}((X,X'))\rightarrow \mathbb{R}(k_X(X')).$
The reproducing kernel $k(X,X')$ is positive definite since it is symmetric, and for any $n\in\mathbb{N}$, $\{X_i\}_{i=1}^n\in\mathcal{X}$ and $\{c_i\}_{i=1}^n\in\mathbb{R}$, we have:
\begin{align*}
\sum_{i=1}^n\sum_{j=1}^nc_ic_jk(X_i,X_j)=\sum_{i=1}^n\sum_{j=1}^n\langle c_i k(X_i,.),c_jk(X_j,.)\rangle_{\mathcal{H}}=\Vert \sum_{i=1}^nc_ik(X_i,.)\Vert_{\mathcal{H}}^2\geq 0.
\end{align*}
For more background on RKHS, we refer to various standard references such as \cite{aronszajn50reproducing}, \cite{saitoh1988theory}, and \cite{Berlinet2004ReproducingKH}.\\
In this work, the idea is to construct a RKHS $\mathcal{H}$ such that any function $f$ in $\mathcal{H}$ is decomposed as its Hoeffding decomposition, and therefore, any function $f$ in $\mathcal{H}$ is a candidate to approximate the Hoeffding decomposition of $m$.  
To do so, the method of \cite{DURRANDE201357} as described below is used. 

Let $\mathcal{X} = \mathcal{X}_{1} \times \ldots \times \mathcal{X}_{d}$ be a subset
of $\mathbb{R}^{d}$. For each $a\in \{1,\cdots,d\}$, we choose a RKHS  $\mathcal{H}_{a}$ and its associated kernel $k_{a}$ defined on the set $\mathcal{X}_{a} \subset \mathbb{R}$ such that the two following properties are satisfied:
\begin{itemize}
\item[(i)]\label{intro:idurr} $k_{a}:\mathcal{X}_{a} \times \mathcal{X}_{a} \rightarrow \mathbb{R}$ is $P_{a}\otimes P_{a}$ measurable,
\item[(ii)]\label{intro:iidurr} $E_{X_{a}}\sqrt{k_{a}(X_{a}, X_{a})} < \infty$.
\end{itemize}
The property \hyperref[intro:iidurr]{(ii)} depends on the kernel $k_a$, $a=1,...,d$ and the distribution of $X_a$, $a=1,...,d$. It is not very restrictive since it is satisfied, for example, for any bounded kernel.

The RKHS $\mathcal{H}_{a}$ can be decomposed as a sum of two orthogonal
sub-RKHS, $\mathcal{H}_{a} = \mathcal{H}_{0 a}\stackrel{\perp}{\oplus} \mathcal{H}_{1 a},$
%$$\mathcal{H}_{a} = \mathcal{H}_{0 a}\stackrel{\perp}{\oplus} \mathcal{H}_{1 a},$$
where $\mathcal{H}_{0 a}$ is the RKHS of zero mean functions, $\mathcal{H}_{0 a} = \{ f_{a} \in \mathcal{H}_{a}:\: E_{X_{a}}(f_{a}(X_{a})) =
   0\},$
and $\mathcal{H}_{1 a}$ is the RKHS of constant functions, $\mathcal{H}_{1 a} = \{  f_{a} \in \mathcal{H}_{a}:\: f_{a}(X_{a}) =C \}.$
The kernel $k_{0a}$ associated with the RKHS $\mathcal{H}_{0 a}$ is defined
by:
\begin{equation}
\label{kernelmodify}
k_{0a} (X_{a},X'_{a}) = k_{a}(X_{a},X'_{a}) - 
\frac{E_{U \sim P_{a}}(k_{a}(X_{a},U))E_{U \sim P_{a}}(k_{a}(X'_{a},U))}
{E_{(U,V)\sim P_{a}\otimes P_{a}}k_{a}(U,V)}.
\end{equation} 
Let $k_{v}(X_{v}, X'_{v}) = \prod_{a \in v} k_{0a} (X_{a},X'_{a}),$ then the  ANOVA kernel $k(.,.)$ is defined as follows:
\begin{equation*}
 k(X, X') = \prod_{a=1}^{d} 
\Big(1+k_{0a}(X_{a}, X'_{a})\Big) = 
1 + \sum_{v \in \mathcal{P}} k_{v}(X_{v}, X'_{v}).
\end{equation*}
For $\mathcal{H}_{v}$ being the RKHS associated with the kernel $k_{v}$, the 
RKHS associated with the ANOVA kernel is then defined by $\mathcal{H} = \prod_{a=1}^{d}( \mathbbm{1} \stackrel{\perp}{\oplus}
   \mathcal{H}_{0a}) = \mathbbm{1} + \sum_{v \in \mathcal{P}} \mathcal{H}_{v},$
where $\perp$ denotes the $L^2$ inner product.
According to this construction, any function $f \in \mathcal{H}$ satisfies decomposition \eref{durandhoeff}, $f(X)=\langle f,k(X,.)\rangle_\mathcal{H}=f_0+\sum_{v\in\mathcal{P}}f_v(X_v),$
which is the Hoeffding decomposition of $f$. 
 
The regularity properties of the RKHS $\mathcal{H}$ constructed as described above, depend on the set of the kernels ($k_{a}$, $a=1,...,d$). This method allows us to choose different approximation spaces independently of the distribution of the input variables $X_1,...,X_d$, by choosing different sets of kernels. While, as mentioned earlier, in the meta-modelling approach based on the polynomial Chaos expansion, according to the distribution of the input variables $X_1,...,X_d$, a unique family of orthonormal polynomials $\{\phi_j\}_{j=0}^\infty$ is determined.
Here, the distribution of the components of $X$ occurs only for the orthogonalization of the spaces $\mathcal{H}_v$, $v\in\mathcal{P}$, and not in the choice of the RKHS, under the condition that properties \hyperref[intro:idurr]{(i)} and \hyperref[intro:iidurr]{(ii)} are satisfied. This is one of the main advantages of this method compared to the method based on the truncated polynomial Chaos expansion where the smoothness of the approximation is handled only by the choice of the truncation \cite{BlatmanSudret}. 
\subsubsection{RKHS group lasso algorithm}\label{appendix:gls}
We consider the minimization of the RKHS group lasso criterion given by,
$$C_g(f_0,\theta)=\Vert Y-f_0I_n-\sum_{v\in\mathcal{P}}K_v\theta_v\Vert^2+\sqrt{n}\mu_g\sum_{v\in\mathcal{P}}\Vert K_v^{1/2}\theta_v\Vert .$$ 
We begin with the constant term $f_0$. The ordinary first derivative of the function $C_g(f_0,\theta)$ at $f_0$ is equal to:
\begin{equation*}
\frac{\partial C_g}{\partial f_0}=-2\sum_{i=1}^n(Y-f_0I_n-\sum_{v\in\mathcal{P}}K_v\theta_v),
\end{equation*}
and therefore, 
\begin{align}\label{fzchapo}
\widehat{f_0}=\frac{1}{n}\sum_{i=1}^n{Y_i}-\frac{1}{n}\sum_i\sum_v(K_v\theta_v)_i,
\end{align}
%$$\widehat{f_0}=\frac{1}{n}\sum_{i=1}^n{Y_i}-\frac{1}{n}\sum_i\sum_v(K_v\theta_v)_i,$$
where $(K_v\theta_v)_i$ denotes the i-th component of $K_v\theta_v$. 

The next step is to calculate $\widehat{\theta}=\text{argmin}_{\theta\in\mathbb{R}^{n\times\vert \mathcal{P}\vert}}C_g(f_0,\theta).$ %To do so, we consider a group $v$, and fix the values of the parameters for all the other groups. The partial derivative of $C_g(f_0,\theta)$ with respect to the block $v$ is:
Since $C_g(f_0,\theta)$ is convex and separable, we use a block coordinate descent algorithm, group $v$ by group $v$. In the following, we fix a group $v$, and we find the minimizer of $C_g(f_0,\theta)$ with respect to $\theta_v$ for the given values of $f_0$ and $\theta_w$, $w\neq v$. Set
\begin{align*}
C_{g,v}(f_0,\theta_v)=\Vert R_v-K_v\theta_v\Vert^2+\sqrt{n}\mu_g\Vert K_v^{1/2}\theta_v\Vert,
\end{align*}
where 
\begin{equation}
\label{residul}
R_v = Y-f_0-\sum_{w\neq v}K_w\theta_w.
\end{equation}
We aim to minimize $C_{g,v}(f_0,\theta_v)$ with respect to $\theta_v$. Let $\partial C_{g,v}$ be the sub-differential of $C_{g,v}(f_0,\theta_v)$ with respect to $\theta_v$:
\begin{equation*}
\partial C_{g,v}(f_0,\theta)=\{-2K_v(R_v-K_v\theta_v)+\sqrt{n}\mu_gt_v\: :\: t_v\in\partial\Vert K_v^{1/2}\theta_v\Vert\}.
\end{equation*}
The first order optimality condition (see Preliminary \eref{fooc}) ensures the existence of $\widehat{t}_v\in\partial\Vert K_v^{1/2}\theta_v\Vert$ fulfilling, 
\begin{equation}
\label{fop}
-2K_v(R_v-K_v\theta_v)+\sqrt{n}\mu_g\widehat{t}_v=0.
\end{equation}
Using the sub-differential definition (see Preliminary \ref{rem:subdiff}), we obtain: %$\partial\Vert K_v^{1/2}\theta_v\Vert= \{{K_v\theta_v/\Vert K_v^{1/2}\theta_v\Vert}\}$ if $\theta_v\neq 0,$ and $\partial\Vert K_v^{1/2}\theta_v\Vert=\{\widehat{t}_v\in\mathbb{R}^n,\:\Vert K_v^{-1/2}\widehat{t}_v\Vert\leq 1\}$ if $\theta_v=0.$
\begin{align*}
\partial\Vert K_v^{1/2}\theta_v\Vert= \{\frac{K_v\theta_v}{\Vert K_v^{1/2}\theta_v\Vert}\}\quad \text{if}\quad \theta_v\neq 0,
\end{align*}
and,
\begin{align*}
 \partial\Vert K_v^{1/2}\theta_v\Vert=\{\widehat{t}_v\in\mathbb{R}^n,\:\Vert K_v^{-1/2}\widehat{t}_v\Vert\leq 1\}\quad\text{if}\quad \theta_v=0.
\end{align*}
Let $\widehat{\theta}_v$ be the minimizer of $C_{g,v}$. The sub-differential equations above give the two following cases:\\ 
Case 1. If $\widehat{\theta}_v=0$, then there exists $\widehat{t}_v\in\mathbb{R}^n$ such that $\Vert K_v^{-1/2}\widehat{t}_v\Vert\leq 1$ and it fulfils Equation~\eref{fop}, $2K_vR_v=\sqrt{n}\mu_g\widehat{t}_v$. 
%\begin{equation*}
%2K_vR_v=\sqrt{n}\mu_g\widehat{t}_v,
%\end{equation*}
Therefore, the necessary and sufficient condition for which the solution $\widehat{\theta}_v = 0$ is the optimal one is $2\Vert K_v^{1/2}R_v\Vert/\sqrt{n}\leq\mu_g.$\\
Case 2. If $\widehat{\theta}_v\neq0$, then $\widehat{t}_v=K_v\widehat{\theta}_v/\Vert K_v^{1/2}\widehat{\theta}_v\Vert$ and it fulfils Equation~\eref{fop}, 
\begin{align*}
2K_v(R_v-K_v\widehat{\theta}_v)=\sqrt{n}\mu_g\frac{K_v\widehat{\theta}_v}{\Vert K_v^{1/2}\widehat{\theta}_v\Vert}.
\end{align*}
We obtain then,
\begin{align}
\label{ttnzero}
\widehat{\theta}_v=(K_v+\frac{\sqrt{n}\mu_g}{2\Vert K_v^{1/2}\widehat{\theta}_v\Vert}I_n)^{-1}R_v.
\end{align}
Since $\widehat{\theta}_v$ appears in both sides of the Equation \eref{ttnzero}, a numerical procedure is needed:
\begin{prop}\label{theo:ttahat}
For $\rho>0$ let $\theta(\rho)=(K_v+\rho I_n)^{-1}R_v$. There exists a non-zero solution to Equation \eref{ttnzero} if and only if there exists $\rho>0$ such that 
\begin{equation}
\label{condttahat}
\mu_g=\frac{2\rho}{\sqrt{n}}\Vert K_v^{1/2}\theta(\rho)\Vert. 
\end{equation}
Then $\widehat{\theta}_v=\theta(\rho)$. 
\end{prop}
\subparagraph*{Proof}
If there exists a non-zero solution to Equation \eref{ttnzero}, then $\Vert K_v^{1/2}\widehat{\theta}_v\Vert\neq 0$ since $K_v$ is positive definite. Take $\rho={\sqrt{n}\mu_g/2\Vert K_v^{1/2}\widehat{\theta}_v\Vert},$ then 
$\theta(\rho)=(K_v+\frac{\sqrt{n}\mu_g}{2\Vert K_v^{1/2}\widehat{\theta}_v\Vert}I_n)^{-1}R_v=\widehat{\theta}_v,$
and, for such $\rho$ Equation \eref{condttahat} is satisfied. Conversely, if there exists $\rho>0$ such that Equation \eref{condttahat} is satisfied, then $\Vert K_v^{1/2}\theta(\rho)\Vert\neq 0$ and $\rho=\frac{\sqrt{n}\mu_g}{2\Vert K_v^{1/2}\theta(\rho)\Vert}.$ Therefore, 
$\theta(\rho)=(K_v+\frac{\sqrt{n}\mu_g}{2\Vert K_v^{1/2}\theta(\rho)\Vert}I_n)^{-1}R_v,$
which is Equation \eref{ttnzero} calculated in $\widehat{\theta}_v=\theta(\rho)$.
\hfill $\Box$  

\begin{rem}
Define $y(\rho)=2\rho\Vert K_v^{1/2}\theta(\rho)\Vert -\sqrt{n}\mu_g$ with $\theta(\rho)=(K_v+\rho I_n)^{-1}R_v$, then $y(\rho)=0$ has a unique solution, denoted $\widehat{\rho}$, which leads to calculate $\widehat{\theta}(\widehat{\rho})$.
\end{rem}
\subparagraph*{Proof}
For $\rho=0$ we have $y(0)=-\sqrt{n}\mu_g<0$, since $\mu_g>0$; and for $\rho\rightarrow +\infty$ we have $y(\rho)>0$, since $\Vert K_v^{1/2}(\frac{K_v}{\rho}+I_n)^{-1}R_v\Vert\rightarrow\Vert K_v^{1/2}R_v\Vert$ and $\Vert 2K_v^{1/2}R_v\Vert >\sqrt{n}\mu_g$.

Moreover, we have $y(\rho)=2\Vert (\frac{I_n}{\rho}+k_v^{-1})^{-1}k_v^{-1/2}R_v\Vert-\sqrt{n}\mu_g$, which is equal to $2(X^TA^{-2}X)^{1/2}-\sqrt{n}\mu_g$ with $A=(I_n/\rho+k_v^{-1})$ and $X=k_v^{-1/2}R_v$.
%\begin{align*}
%y(\rho)&=2\Vert (\frac{I_n}{\rho}+k_v^{-1})^{-1}k_v^{-1/2}R_v\Vert-\sqrt{n}\mu_g,\\
%&=2(X^TA^{-2}X)^{1/2}-\sqrt{n}\mu_g,
%\end{align*}
%where $A=(I_n/\rho+k_v^{-1})$ and $X=k_v^{-1/2}R_v$. 
The first derivative of $y(\rho)$ in $\rho$ is obtained by $\frac{\partial y(\rho)}{\partial\rho}=(X^TA^{-2}X)^{-1/2}\frac{\partial (X^TA^{-2}X)}{\partial\rho}$. Finally, by simple calculations we get,  
%\begin{align*}
%\frac{\partial y(\rho)}{\partial\rho}=(X^TA^{-2}X)^{-1/2}\frac{\partial (X^TA^{-2}X)}{\partial\rho},
%\end{align*}
%and,
%\begin{align*}
%\frac{\partial (X^TA^{-2}X)}{\partial\rho} &= X^T\frac{\partial (A^{-1})^2}{\partial\rho}X,\\
%&=2X^TA^{-1}(-A^{-1}\frac{\partial A}{\partial \rho}A^{-1})X,\\
%&=\frac{2}{\rho^2}\Vert A^{-3/2}X\Vert.
%\end{align*}
%Finally, we get 
\begin{align*}
\frac{\partial y(\rho)}{\partial\rho}=\frac{2\Vert (\frac{I_n}{\rho}+k_v^{-1})^{-3/2}k_v^{-1/2}R_v\Vert}{\rho^2\Vert (\frac{I_n}{\rho}+k_v^{-1})^{-1}k_v^{-1/2}R_v\Vert}>0.
\end{align*}
Therefore $y(\rho)$ is an increasing function of $\rho$, and the proof is complete.
\hfill $\Box$ 

In order to calculate $\rho$ and so $\widehat{\theta}_v=\theta(\rho)$, we use Algorithm \ref{algo:rho} which is a part of the RKHS group lasso Algorithm \ref{algo:GL} when $\widehat{\theta}_v\neq 0$.
\begin{algorithm}[h!]%[htbp]%[h!]
\caption{Algorithm to find $\rho$ as well as $\widehat{\theta}_v$}\label{algo:rho}
\begin{algorithmic}[1]
\If{$\widehat{\theta}_{\text{old}}=0$} \Comment{$\widehat{\theta}_{\text{old}}$ is $\widehat{\theta}_v$ computed in the previous step of the RKHS group lasso algorithm.}
\State{Set $\rho\gets 1$ and calculate $y(\rho)$}
 \If{$y(\rho)>0$}
 \State{Find $\widehat{\rho}$ that minimizes $y(\rho)$ on the interval $[0,1]$}
 \Else
 \Repeat
  \State{Set $\rho\gets\rho\times 10$ and calculate $y(\rho)$}
 \Until{$y(\rho)>0$} 
 \State{Find $\widehat{\rho}$ that minimizes $y(\rho)$ on the interval $[\rho/10,\rho]$}
 \EndIf 
\Else
\State{Set $\rho\gets \frac{\sqrt{n}\mu_g}{2\Vert K_v^{1/2}\widehat{\theta}_{\text{old}}\Vert}$ and calculate $y(\rho)$}
 \If{$y(\rho)>0$}
 \Repeat
  \State{Set $\rho\gets\rho/10$ and calculate $y(\rho)$}
 \Until{$y(\rho)<0$} 
 \State{Find $\widehat{\rho}$ that minimizes $y(\rho)$ on the interval $[\rho,\rho\times 10]$}
 \Else
 \Repeat
  \State{Set $\rho\gets\rho\times 10$ and calculate $y(\rho)$}
 \Until{$y(\rho)>0$} 
 \State{Find $\widehat{\rho}$ that minimizes $y(\rho)$ on the interval $[\rho/10,\rho]$}
 \EndIf 
\EndIf 
\State{calculate $\widehat{\theta}_v=\theta(\widehat{\rho})$}
\end{algorithmic}
\end{algorithm}
\subsubsection{Computational cost}
The complexity for the matrices $K_v$, $v\in\mathcal{P}$ is equal to $n^3$, which is given by the singular value decomposition to get eigenvalues and eigenvectors of each $K_v$.
Supposing that the matrices $K_v$, $v\in\mathcal{P}$ was first created and are already stored, the complexity for the constant term $f_0$ is given by the second term in equation \eref{fzchapo}, which is equal to $n^2$. Given $\widehat{\rho}$, the complexity for $\widehat{\theta}_v$, $v\in\mathcal{P}$, is given by the backsolving of $(K_v+\widehat{\rho} I_n)\widehat{\theta}_v=R_v$ to get $\widehat{\theta}_v$, which is equal to $n^2$.  The computation of $\widehat{\rho}$ is done using Brent-Dekker method implemented as function \fct{gsl\_root\_fsolver\_brent} from \cite{galassi2018scientific}. This method combines an interpolation strategy with the bisection algorithm and takes $O(m)$ iterations to converge, where $m$ is the number of steps that the bisection algorithm would take.   
\subsubsection{RKHS ridge group sparse algorithm}
We consider the minimization of the RKHS ridge group sparse criterion:
$$C(f_0,\theta)=\Vert Y-f_0I_n-\sum_{v\in\mathcal{P}}K_v\theta_v\Vert^2+\sqrt{n}\gamma\sum_{v\in\mathcal{P}}\Vert K_v\theta_v\Vert+n\mu\sum_{v\in\mathcal{P}}\Vert K_v^{1/2}\theta_v\Vert.$$
The constant term $f_0$ is estimated as in the RKHS group lasso algorithm. In order to calculate $\widehat{\theta}=\text{argmin}_{\theta\in\mathbb{R}^{n\times\vert\mathcal{P}\vert}}C(f_0,\theta)$, we use once again the block coordinate descent algorithm group $v$ by group $v$. In the following, we fix a group $v$, and we find the minimizer of $C(f_0,\theta)$ with respect to $\theta_v$ for given values of $f_0$ and $\theta_w$, $w\neq v$. We aim at minimizing with respect to $\theta_v$,
\begin{align*}
C_v(f_0,\theta_v)=\Vert R_v-K_v\theta_v\Vert^2+\sqrt{n}\gamma\Vert K_v\theta_v\Vert+n\mu\Vert K_v^{1/2}\theta_v\Vert,
\end{align*}
where $R_v$ is defined by \eref{residul}.

Let $\partial C_{v}$ be the sub-differential of $C_{v}(f_0,\theta_v)$ with respect to $\theta_v$,
\begin{equation*}
\partial C_v=\{-2K_v(R_v-K_v\theta_v)+\sqrt{n}\gamma s_v+n\mu t_v\: :\: s_v\in\partial\Vert K_v\theta_v\Vert,\quad t_v\in\partial\Vert K_v^{1/2}\theta_v\Vert\},
\end{equation*} 
According to the first order optimality condition (see Preliminary \ref{fooc}), we know that there exists $\widehat{s}_v\in\partial\Vert K_v\theta_v\Vert$ and $\widehat{t}_v\in\partial\Vert K_v^{1/2}\theta_v\Vert$ such that, 
\begin{equation}
\label{foprgs}
-2K_v(R_v-K_v\theta_v)+\sqrt{n}\gamma\widehat{s}_v+n\mu\widehat{t}_v=0.
\end{equation}
The sub-differential definition (see Preliminary \ref{rem:subdiff}) gives: %$\{\partial\Vert K_v^{1/2}\theta_v\Vert= \{\frac{K_v\theta_v}{\Vert K_v^{1/2}\theta_v\Vert}\},\:\partial\Vert K_v\theta_v\Vert= \{\frac{K_v^2\theta_v}{\Vert K_v\theta_v\Vert}\}\}$ if $\theta_v\neq 0,$ and $\{\partial\Vert K_v^{1/2}\theta_v\Vert=\{\widehat{t}_v\in\mathbb{R}^n,\Vert K_v^{-1/2}\widehat{t}_v\Vert\leq 1\},\:\partial\Vert K_v\theta_v\Vert=\{\widehat{s}_v\in\mathbb{R}^n,\Vert K_v^{-1} \widehat{s}_v\Vert\leq 1\}\}$ if $\theta_v=0.$
\begin{align*}
\{\partial\Vert K_v^{1/2}\theta_v\Vert= \{\frac{K_v\theta_v}{\Vert K_v^{1/2}\theta_v\Vert}\},\:\partial\Vert K_v\theta_v\Vert= \{\frac{K_v^2\theta_v}{\Vert K_v\theta_v\Vert}\}\}\quad \text{if}\quad \theta_v\neq 0,
\end{align*}
and,
\begin{align*}
 \{\partial\Vert K_v^{1/2}\theta_v\Vert=\{\widehat{t}_v\in\mathbb{R}^n,\Vert K_v^{-1/2}\widehat{t}_v\Vert\leq 1\},\:\partial\Vert K_v\theta_v\Vert=\{\widehat{s}_v\in\mathbb{R}^n,\Vert K_v^{-1} \widehat{s}_v\Vert\leq 1\}\}\quad \text{if}\quad \theta_v=0.
\end{align*}
Let $\widehat{\theta}_v$ be the minimizer of the $C_v(f_0,\theta_v)$.
Using the sub-differential equations above, the estimator $\widehat{\theta}_v$, $v\in\mathcal{P}$ is obtained following the two cases below:\\ 
Case 1. If $\widehat{\theta}_v=0$, then there exists $\widehat{s}_v\in\mathbb{R}^n$ such that $\Vert K_v^{-1}\widehat{s}_v\Vert\leq 1$ and it fulfils Equation \eref{foprgs}, $2K_vR_v-n\mu \widehat{t}_v=\sqrt{n}\gamma \widehat{s}_v,$ 
%\begin{equation*}
%2K_vR_v-n\mu \widehat{t}_v=\sqrt{n}\gamma \widehat{s}_v,
%\end{equation*}
with $\widehat{t}_v\in\mathbb{R}^n$, $\Vert K_v^{-1/2}\widehat{t}_v\Vert\leq 1$. Set $J(\widehat{t}_v)=\Vert 2R_v-n\mu K_v^{-1}\widehat{t}_v\Vert,$ and, 
$J^*=\text{argmin}_{\widehat{t}_v\in\mathbb{R}^n}\{J(\widehat{t}_v),\text{ such that }\Vert K_v^{-1/2}\widehat{t}_v\Vert\leq 1\}.$ 
Then the solution to Equation \eref{foprgs} is zero if and only if $J^*\leq\gamma$.\\
Case 2. If $\widehat{\theta}_v\neq0$, then we have $\widehat{s}_v=K_v^2\widehat{\theta}_v/\Vert K_v\widehat{\theta}_v\Vert$, and $\widehat{t}_v=K_v\widehat{\theta}_v/\Vert K_v^{1/2}\widehat{\theta}_v\Vert$ fulfilling Equation~\eref{foprgs}, $2K_v(R_v-K_v\widehat{\theta}_v)=\sqrt{n}\gamma\frac{K_v^2\widehat{\theta}_v}{\Vert K_v\widehat{\theta}_v\Vert_2}+n\mu\frac{K_v\widehat{\theta}_v}{\Vert K_v^{1/2}\widehat{\theta}_v\Vert},$ 
%\begin{align*}
%2K_v(R_v-K_v\widehat{\theta}_v)=\sqrt{n}\gamma\frac{K_v^2\widehat{\theta}_v}{\Vert K_v\widehat{\theta}_v\Vert_2}+n\mu\frac{K_v\widehat{\theta}_v}{\Vert K_v^{1/2}\widehat{\theta}_v\Vert},
%\end{align*}
that is, 
\begin{align}\label{tthatgrp}
\widehat{\theta}_v=(K_v+\frac{\sqrt{n}\gamma}{2\Vert K_v\widehat{\theta}_v\Vert}K_v+\frac{n\mu}{2\Vert K_v^{1/2}\widehat{\theta}_v\Vert}I_n)^{-1}R_v\quad \text{if}\quad \widehat{\theta}_v\neq0.
\end{align}
In this case the calculation of $\widehat{\theta}_v$ needs a numerical algorithm.
\begin{prop} (Proposition 8.4 in \cite{huet:hal-01434895})\label{theo:ttahatgrp}
For $\rho_1,\rho_2>0$, let $\theta(\rho_1,\rho_2)=(K_v+\rho_1K_v+\rho_2I_n)^{-1}R_v$. If $\mu>0$, there exists a non zero solution to Equation \eref{tthatgrp} if and only if there exists $\rho_1,\rho_2>0$ such that $\gamma=\frac{2\rho_1}{\sqrt{n}}\Vert K_v\theta(\rho_1,\rho_2)\Vert,\:\mbox{and}\:\mu=\frac{2\rho_2}{n}\Vert K^{1/2}_v\theta(\rho_1,\rho_2)\Vert.$ 
Then $\widehat{\theta}_v=\theta(\rho_1,\rho_2)$. 
\end{prop}
\subparagraph*{Proof} The proof is given in \cite{huet:hal-01434895}.
\subsubsection{Computational cost}
The complexity for the matrices $K_v$, $v\in\mathcal{P}$ and the constant term $f_0$ is the same as for RKHS group lasso algorithm. Given $\widehat{\rho}_1,\widehat{\rho}_2$, the complexity for $\widehat{\theta}_v$, $v\in\mathcal{P}$, is given by the backsolving of $(K_v+\rho_1K_v+\rho_2I_n)\widehat{\theta}_v=R_v$ to get $\widehat{\theta}_v$, which is equal to $n^2$.  The computation of $\widehat{\rho}_1,\widehat{\rho}_2$, is insured using a combination of three methods: first we implement a modified version of Newton's method, if it does not achieve the convergence second we implement a version of the Hybrid algorithm and if it does not achieve the convergence, third we implement a version of the discrete Newton algorithm called Broyden algorithm. These methods are implemented as functions \fct{gsl\_multiroot\_fdfsolver\_gnewton}, \fct{gsl\_multiroot\_fsolver\_hybrids}, and \fct{gsl\_multiroot\_fsolver\_broyden} from \cite{galassi2018scientific}, respectively. These methods converge fast in general, and for $n$ big enough their complexity is dominated by $n^2$.  
\subsection{Overview of the RKHSMetaMod functions}\label{sec:Overview}
In the R environment, one can install and load the \pkg{RKHSMetaMod} package by using the following commands:
%\texttt{
%R> install.packages("RKHSMetaMod")\\
%R> library("RKHSMetaMod")\\
%}
\begin{example}
  install.packages("RKHSMetaMod")
  library("RKHSMetaMod")
\end{example}
The optimization problems in this package are solved using block coordinate descent algorithm which requires various computational algorithms including generalized Newton, Broyden and Hybrid methods. 
In order to gain the efficiency in terms of the calculation time and be able to deal with high dimensional problems, the computationally efficient tools of C++ packages \pkg{Eigen} \cite{eigenweb} and \pkg{GSL} \cite{galassi2018scientific} via \pkg{RcppEigen} \cite{RcppEigen} and \pkg{RcppGSL} \cite{RcppGSL} packages are used in the \pkg{RKHSMetaMod} package. For different examples of usage of \pkg{RcppEigen} and \pkg{RcppGSl} functions see the work by \cite{Eddelbuettel:2013:SRC:2517725}. 

The complete documentation of \pkg{RKHSMetaMod} package is available from CRAN \cite{rkhsmeamodpackage}. Here, a brief documentation of some of its main and companion functions is presented in the next two Sections.   

\subsubsection{Main RKHSMetaMod functions}\label{subsec:Main}
Let us begin by introducing some notations. For a given Dmax$\in\mathbb{N}$, let $\mathcal{P}_{\text{Dmax}}$ be the set of parts of $\{1,...,d\}$ with dimension $1$ to Dmax. The cardinal of $\mathcal{P}_{\text{Dmax}}$ is denoted by $\text{vMax}=\sum_{j=1}^{\text{Dmax}}\binom{d}{j}.$
%$$\text{vMax}=\sum_{j=1}^{\text{Dmax}}\binom{d}{j}.$$
\subparagraph{\fct{RKHSMetMod} function:} For a given value of Dmax and a chosen kernel (see Table \ref{kernels}), this function calculates the Gram matrices $K_v$, $v\in\mathcal{P}_{\text{Dmax}}$, and produces a sequence of estimators $\widehat{f}$ associated with a given grid of values of tuning parameters $\mu,\gamma$, i.e. the solutions to the RKHS ridge group sparse (if $\gamma\neq0$) or the RKHS group lasso problem (if $\gamma=0$).
Table \ref{metmod} gives a summary of all the input arguments of the \fct{RKHSMetMod} function as well as the default values for non-mandatory arguments. 
\begin{table}[h!]%[htbp]%[h!]
\centering
\begin{tabular}{l|p{10cm}} 
 Input parameter &  Description \\ \hline
Y &  Vector of the response observations of size $n$.\\ 
X &  \multicolumn{1}{m{10cm}}{Matrix of the input observations with $n$ rows and $d$ columns. Rows correspond to the observations and columns correspond to the variables.}\\ 
kernel & \multicolumn{1}{m{10cm}}{Character, indicates the type of the kernel (see Table \ref{kernels}) chosen to construct the RKHS $\mathcal{H}$.}\\
Dmax & \multicolumn{1}{m{10cm}}{Integer, between $1$ and $d$, indicates the maximum order of interactions considered in the RKHS
meta-model: Dmax$=1$ is used to consider only the main effects, Dmax$=2$ to
include the main effects and the second-order interactions, and so on.}\\
gamma & \multicolumn{1}{m{10cm}}{Vector of non-negative scalars, values of the tuning parameter $\gamma$ in decreasing
order. If $\gamma=0$ the function solves the RKHS group lasso optimization problem and for $\gamma>0$ it solves the RKHS ridge group sparse optimization problem.}\\
frc & \multicolumn{1}{m{10cm}}{Vector of positive scalars. Each element of the vector sets a value to the tuning parameter $\mu$: $\mu=\mu_{\text{max}}/(\sqrt{n}\times\text{frc})$. The value $\mu_{\text{max}}$ (see Equation \eref{maxmu}) is calculated inside the program.}\\
verbose & \multicolumn{1}{m{10cm}}{Logical. Set as TRUE to print: the group $v$ for which the correction of the Gram matrix $K_v$ is done (see Section \ref{subsec:Gramm}), and for each pair of the tuning parameters $(\mu,\gamma)$: the number of current iteration, active groups and convergence criteria. It is set as FALSE by default.}\\
\end{tabular}
\caption{List of the input arguments of the \fct{RKHSMetMod} function. \label{metmod}}
\end{table} 

The \fct{RKHSMetMod} function returns a list of $l$ components, with $l$ being equal to the number of pairs of the tuning parameters $(\mu,\gamma)$, i.e. $l=\vert \text{gamma}\vert\times\vert\text{frc}\vert$. Each component of the list is a list of three components "mu", "gamma" and "Meta-Model":
%Each component of the list is an instance of the "RKHSMetMod" class. Its three attributes contain the following outputs:
\begin{itemize}
\item mu: value of the tuning parameter $\mu$ if $\gamma>0$, or $\mu_g=\sqrt{n}\times\mu$ if $\gamma=0$.
\item gamma: value of the tuning parameter $\gamma$.
\item Meta-Model: a RKHS ridge group sparse or RKHS group lasso object associated with the tuning parameters mu and gamma. Table \ref{metmodoutput} gives a summary of all arguments of the output "Meta-Model" of \fct{RKHSMetMod} function. 
\end{itemize}
\begin{table}[h!]%[htbp]%[h!]
\centering
\begin{tabular}{l|p{10cm}} 
 Output parameter &  Description \\ \hline
intercept &  \multicolumn{1}{m{10cm}}{Scalar, estimated value of intercept.}\\ 
teta &  \multicolumn{1}{m{10cm}}{Matrix with vMax rows and $n$ columns. Each row of the matrix is the estimated vector $\theta_{v}$ for $v=1,...,$vMax.}\\ 
fit.v & \multicolumn{1}{m{10cm}}{Matrix with $n$ rows and vMax columns. Each row of the matrix is the estimated value of $f_{v}=K_{v}\theta_{v}$.}\\
fitted & \multicolumn{1}{m{10cm}}{Vector of size $n$, indicates the estimator of $m$.}\\
Norm.n & \multicolumn{1}{m{10cm}}{Vector of size vMax, estimated values for the empirical $L^2$-norm.}\\
Norm.H & \multicolumn{1}{m{10cm}}{Vector of size vMax, estimated values for the Hilbert norm.}\\
supp &  \multicolumn{1}{m{10cm}}{Vector of active groups.} \\
Nsupp &  \multicolumn{1}{m{10cm}}{Vector of the names of the active groups.} \\
SCR &  \multicolumn{1}{m{10cm}}{Scalar equals to $\Vert Y-f_{0}-\sum_{v}K_{v}\theta_{v}\Vert ^{2}$.} \\
crit &  \multicolumn{1}{m{10cm}}{Scalar indicates the value of the penalized criterion.} \\
gamma.v &  \multicolumn{1}{m{10cm}}{Vector of size vMax, coefficients of the empirical $L^2$-norm.} \\
mu.v &  \multicolumn{1}{m{10cm}}{Vector of size vMax, coefficients of the Hilbert norm.} \\
iter &  \multicolumn{1}{m{10cm}}{List of two components: maxIter, and the number of iterations until the convergence is achieved.} \\
convergence &  \multicolumn{1}{m{10cm}}{TRUE or FALSE. Indicates whether the algorithm has converged or not.} \\
RelDiffCrit &  \multicolumn{1}{m{10cm}}{Scalar, value of the first convergence criterion at the last iteration, i.e. $\Vert{\theta_{lastIter}-\theta_{lastIter-1}/\theta_{lastIter-1}}\Vert ^{2}$.} \\
RelDiffPar &  \multicolumn{1}{m{10cm}}{Scalar, value of the second convergence criterion at the last iteration, i.e. ${crit_{lastIter}-crit_{lastIter-1}/crit_{lastIter-1}}$.} \\
\end{tabular}
\caption{List of the arguments of the output "Meta-Model" of \fct{RKHSMetMod} function. \label{metmodoutput}}
\end{table} 

\subparagraph{\fct{RKHSMetMod$\_$qmax} function:} For a given value of Dmax and a chosen kernel (see Table \ref{kernels}), this function calculates the Gram matrices $K_v$, $v\in\mathcal{P}_{\text{Dmax}}$; determines $\mu$, reffered to as $\mu_{qmax}$, for which the number of groups in the support of the RKHS group lasso solution is equal to $qmax$; and produces a sequence of estimators $\widehat{f}$ associated with the tuning parameter $\mu_{qmax}$ and a grid of values of the tuning parameter $\gamma$. All the estimators $\widehat{f}$ produced by this function have at most $qmax$ groups in their support.
This function has the following input arguments: 
\begin{itemize}
\item[$-$] $Y$, $X$, kernel, Dmax, gamma, verbose (see Table \ref{metmod}).
\item[$-$] qmax: integer, the maximum number of groups in the support of the obtained solution.
\item[$-$] rat: positive scalar, to restrict the minimum value of $\mu$ considered in Algorithm \ref{algo:GLqmax}, $$\mu_{\text{min}}=\frac{\mu_{\text{max}}}{(\sqrt{n}\times\text{rat})},$$ where the value of $\mu_{\text{max}}$ is given by Equation \eref{maxmu} and is calculated inside the program.
\item[$-$] Num: integer, to restrict the number of different values of the tuning parameter $\mu$ to be evaluated in the RKHS group lasso algorithm until it achieves $\mu_{qmax}$. For example, if Num equals $1$, the program is implemented for three different values of $\mu\in[\mu_{\text{min}},\mu_{\text{max}})$: 
\begin{align*}
&\mu_{1}=\frac{(\mu_{\text{min}}+\mu_{\text{max}})}{2},\\
&\mu_{2}=\left\{ \begin{array}{rcl}
         \frac{(\mu_{\text{min}}+\mu_{1})}{2} & \mbox{if}& \vert \widehat{S}_{\widehat{f}(\mu_1)_{\text{Group Lasso}}}\vert <qmax,\\ 
         \frac{(\mu_{1}+\mu_{\text{max}})}{2}   & \mbox{if} & \vert \widehat{S}_{\widehat{f}(\mu_1)_{\text{Group Lasso}}}\vert >qmax, 
                \end{array}\right.\\
&  \mu_{3}=\mu_{\text{min}},\\              
\end{align*}
where $\vert \widehat{S}_{\widehat{f}(\mu_1)_{\text{Group Lasso}}}\vert$ is the number of groups in the support of the solution of the RKHS group lasso problem, Algorithm \ref{algo:GL}, associated with $\mu_{1}$.

If Num$>1$, the path to cover the interval $[\mu_{\text{min}},\mu_{\text{max}})$ is detailed in Algorithm \ref{algo:GLqmax}. 
\end{itemize}

%The \fct{RKHSMetMod$\_$qmax} function returns an instance of the "RKHSMetMod$\_$qmax" class. Its three attributes contain the followings outputs:
The \fct{RKHSMetMod$\_$qmax} function returns a list of three components "mus", "qs", and "MetaModel":
\begin{itemize}
\item mus: vector of all values of $\mu_i$ in Algorithm \ref{algo:GLqmax}.
\item qs: vector with the same length as mus. Each element of the vector shows the number of groups in the support of the RKHS meta-model obtained by solving RKHS group lasso problem for an element in mus.
\item MetaModel: %list of $l=\vert \text{gamma}\vert$ components. Each component of the list is an instance of the "RKHSMetMod" class for the obtained $\mu_{qmax}$ and one value of the tuning parameter $\gamma$.
list with the same length as the vector gamma. Each component of the list is a list of three components "mu", "gamma" and "Meta-Model":
\begin{itemize}
\item mu: value of $\mu_{qmax}$.
\item gamma: element of the input vector gamma associated with the estimated "Meta-Model".
\item Meta-Model: a RKHS ridge group sparse or RKHS group lasso object associated with the tuning parameters mu and gamma (see Table \ref{metmodoutput}). 
\end{itemize}
\end{itemize}  

\subsubsection{Companion functions}\label{subsec:Comp}
\subparagraph{\fct{calc$\_$Kv} function:} %calculates the Gram matrices $K_v$, $v\in\mathcal{P}$, for a chosen kernel (see Table \ref{kernels}), and returns their associated eigenvalues and eigenvectors. %for $v=1,...,$vMax, with $$\text{vMax}=\sum_{j=1}^{\text{Dmax}}\binom{d}{j}.$$ 
For a given value of Dmax and a chosen kernel (see Table \ref{kernels}), this function calculates the Gram matrices $K_v$, $v\in\mathcal{P}_{\text{Dmax}}$, and returns their associated eigenvalues and eigenvectors. 
This function has,
\begin{itemize}
\item four mandatory input arguments: 
\begin{itemize}
\item $Y$, $X$, kernel, Dmax (see Table \ref{metmod}).
\end{itemize}
\item three facultative input arguments: 
\begin{itemize}
\item correction: logical, set as TRUE to make correction to the matrices $K_v$ (see Section \ref{subsec:Gramm}). It is set as TRUE by default.
\item verbose: logical, set as TRUE to print the group for which the correction is done. It is set as TRUE by default.
\item tol: scalar to be chosen small, set as $1e^{-8}$ by default.
\end{itemize}
\end{itemize}
The \fct{calc$\_$Kv} function returns a list of two components "kv" and "names.Grp":
\begin{itemize}
\item kv: list of vMax components, each component is a list of, 
\begin{itemize}
\item Evalues: vector of eigenvalues.
\item Q: matrix of eigenvectors.
\end{itemize}
\item names.Grp: vector of group names of size vMax.
\end{itemize}

\subparagraph*{\fct{RKHSgrplasso} function:} 
For a given value of the tuning parameter $\mu_g$, this function fits the solution to the RKHS group lasso optimization problem by implementing Algorithm \ref{algo:GL}. This function has, 
\begin{itemize}
\item three mandatory input arguments: 
\begin{itemize}
\item $Y$ (see Table \ref{metmod}). 
\item Kv: list of the eigenvalues and the eigenvectors of the positive definite Gram matrices $K_v$ for $v=1,...,$vMax and their associated group names (output of the function \fct{calc$\_$Kv}).
\item mu: positive scalar indicates the value of the tuning parameter $\mu_g$ defined in Equation \eref{tungrp}.
\end{itemize}
\item two facultative input arguments: 
\begin{itemize}
\item maxIter: integer, to set the maximum number of loops through all groups. It is set as $1000$ by default. 
\item verbose: logical, set as TRUE to print the number of current iteration, active groups and convergence criterion. It is set as FALSE by default.
\end{itemize}
\end{itemize}
This function returns a RKHS group lasso object associated with the tuning parameter $\mu_g$. Its output is a list of $13$ components:
\begin{itemize}
\item intercept, teta, fit.v, fitted, Norm.H, supp, Nsupp, SCR, crit, MaxIter, convergence, RelDiffCrit, and RelDiffPar (see Table \ref{metmodoutput}). 
\end{itemize}
\subparagraph*{\fct{mu$\_$max} function:} 
This function calculates the value $\mu_{\text{max}}$ defined in Equation \eref{maxmu}.
It has two mandatory input arguments: the response vector $Y$, and the list matZ of the eigenvalues and eigenvectors of the positive definite Gram matrices $K_v$ for $v=1,...,$vMax. This function returns the $\mu_{max}$ value.
\subparagraph*{\fct{pen$\_$MetMod} function:} 
This function produces a sequence of the RKHS meta-models associated with a given grid of values of the tuning parameters $\mu,\gamma$. Each RKHS meta-model in the sequence is the solution to the RKHS ridge group sparse optimization problem (obtained by implementing Algorithm \ref{algo:RGS}) associated with a pair of values of $(\mu,\gamma)$ in the grid of values of $\mu,\gamma$.
This function has,
\begin{itemize}
\item seven mandatory input arguments: 
\begin{itemize}
\item $Y$ (see Table \ref{metmod}).
\item gamma: vector of positive scalars. Values of the penalty parameter $\gamma$ in decreasing order.
\item Kv: list of the eigenvalues and the eigenvectors of the positive definite Gram matrices $K_v$ for $v=1,...,$vMax and their associated group names (output of the function \fct{calc$\_$Kv}).
\item mu: vector of positive scalars. Values of the tuning parameter $\mu$ in decreasing order.
\item resg: list of the RKHS group lasso objects associated with the components of "mu", used as initial parameters at \hyperref[step1]{Step $1$}.
\item gama\textunderscore v and mu\textunderscore v: vector of vMax positive scalars. These two inputs are optional. They are provided to associate the weights to the two penalty terms in the RKHS ridge group sparse criterion \eref{parametric}. In order to consider no weights, i.e. all the weights are equal to one, we set these two inputs to scalar zero.
%They set to scalar $0$, to consider no weights, i.e. all weights equal to $1$.
\end{itemize}
\item three facultative input arguments: 
\begin{itemize}
\item maxIter: integer, to set the maximum number of loops through initial active groups at \hyperref[step1]{Step $1$} and maximum number of loops through all groups at \hyperref[step2]{Step $2$}. It is set as $1000$ by default.
\item verbose: logical, set as TRUE to print for each pair of the tuning parameters $(\mu,\gamma)$, the number of current iteration, active groups and convergence criterion. It is set as FALSE by default.
\item calcStwo: logical, set as TRUE to execute \hyperref[step2]{Step $2$}. It is set as FALSE by default.
\end{itemize}
\end{itemize}
%The function \fct{pen$\_$MetMod} returns an RKHS ridge group sparse object associated with each pair of the tuning parameters $(\mu,\gamma)$.
The function \fct{pen$\_$MetMod} returns a list of $l$ components, with $l$ being equal to the number of pairs of the tuning parameters $(\mu,\gamma)$. Each component of the list is a list of three components "mu", "gamma" and "Meta-Model":
\begin{itemize}
\item mu: positive scalar, an element of the input vector "mu" associated with the estimated "Meta-Model".
\item gamma: positive scalar, an element of the input vector "gamma" associated with the estimated "Meta-Model".
\item Meta-Model: a RKHS ridge group sparse object associated with the tuning parameters mu and gamma (see Table \ref{metmodoutput}). 
\end{itemize}

\subparagraph*{\fct{PredErr} function:} 
By considering a testing dataset, this function calculates the prediction errors for the obtained RKHS meta-models. This function has eight mandatory input arguments:
\begin{itemize}
\item[$-$] $X$, gamma, kernel, Dmax (see Table \ref{metmod}).
\item[$-$] $XT$: matrix of observations of the testing dataset with $n^{\text{test}}$ rows and $d$ columns.
\item[$-$] $YT$: vector of response observations of the testing dataset of size $n^{\text{test}}$.
\item[$-$] mu: vector of positive scalars. Values of the tuning parameter $\mu$ in decreasing order.
\item[$-$] res: list of the estimated RKHS meta-models for the learning dataset associated with the tuning parameters $(\mu,\gamma)$ (it could be the output of one of the functions \fct{RKHSMetMod} or \fct{pen$\_$MetMod}).
\end{itemize}
Note that, the same kernel and Dmax have to be chosen as the ones used for the learning dataset.

The function \fct{PredErr} returns a matrix of the prediction errors. Each element of the matrix corresponds to the prediction error of one RKHS meta-model in "res".
\subparagraph*{\fct{prediction} function:} 
This function calculates the predicted values for a new dataset based on the \textit{best} RKHS meta-model estimator. It has six input arguments:
\begin{itemize}
\item[$-$] X, kernel, Dmax (see Table \ref{metmod}).
\item[$-$] Xnew: matrix of new observations with $n^{\text{new}}$ rows and $d$ columns.
\item[$-$] res: list of the estimated RKHS meta-models for a learning dataset associated with the tuning parameters $(\mu,\gamma)$ (it could be the output of one of the functions \fct{RKHSMetMod}, \fct{RKHSMetMod$\_$qmax} or \fct{pen$\_$MetMod}).
\item[$-$] Err: matrix of the prediction errors associated with the RKHS meta-models in "res" (output of the function \fct{PredErr}).
\end{itemize}
The function \fct{prediction} returns a vector of the predicted values based on the \textit{best} RKHS meta-model estimator in "res". This function is available at \href{https://github.com/halalehkamari/RKHSMetaMod}{GitHub}. 
\subparagraph*{\fct{SI$\_$emp} function:} %calculates the empirical Sobol indices for an input or a group of inputs. It has two input arguments:
For each RKHS meta-model $\widehat{f}$, this function calculates the empirical Sobol indices for all the groups that are active in the support of $\widehat{f}$. This function has two input arguments:
\begin{itemize}
\item[$-$] res: list of the estimated meta-models using RKHS ridge group sparse or RKHS group lasso algorithms (it could be the output of one of the functions \fct{RKHSMetMod}, \fct{RKHSMetMod$\_$qmax} or \fct{pen$\_$MetMod}).
\item[$-$] ErrPred: matrix or NULL. If matrix, each element of the matrix corresponds to the prediction error of a RKHS meta-model in "res" (output of the function \fct{PredErr}). Set as NULL by default.
\end{itemize}
The empirical Sobol indices are then calculated for each RKHS meta-model in "res", and a list of vectors of the Sobol indices is returned.
If the argument "ErrPred" is the matrix of the prediction errors, the vector of empirical Sobol indices is returned for the \textit{best} RKHS meta-model in the "res".   

%\end{subappendices}
%\section{Another section}

%This section may contain a figure such as Figure~\ref{figure:rlogo}.

%\begin{figure}[htbp]
%  \centering
%  \includegraphics{Rlogo-5}
%  \caption{The logo of R.}
%  \label{figure:rlogo}
%\end{figure}

%\section{Another section}

%There will likely be several sections, perhaps including code snippets, such as:

%\begin{example}
%  x <- 1:10
%  result <- myFunction(x)
%\end{example}

%\section{Summary}

%This file is only a basic article template. For full details of \emph{The R Journal} style and information on how to prepare your article for submission, see the \href{https://journal.r-project.org/share/author-guide.pdf}{Instructions for Authors}.
\newpage
\bibliography{biblio}

\begin{thebibliography}{}

\bibitem[\protect\citeauthoryear{Aronszajn}{Aronszajn}{1950}]{aronszajn50reproducing}
Aronszajn, N. (1950).
\newblock {Theory of Reproducing Kernels}.
\newblock {\em Transactions of the American Mathematical Society\/}~{\em
  68\/}(3), 337--404.

\bibitem[\protect\citeauthoryear{Bates and Eddelbuettel}{Bates and
  Eddelbuettel}{2013}]{RcppEigen}
Bates, D. and D.~Eddelbuettel (2013).
\newblock {Fast and Elegant Numerical Linear Algebra Using the RcppEigen
  Package}.
\newblock {\em Journal of Statistical Software\/}~{\em 52\/}(5), 1--24.

\bibitem[\protect\citeauthoryear{Berlinet and Thomas-Agnan}{Berlinet and
  Thomas-Agnan}{2003}]{Berlinet2004ReproducingKH}
Berlinet, A. and C.~Thomas-Agnan (2003).
\newblock {\em {Reproducing Kernel Hilbert Spaces in Probability and
  Statistics}}.
\newblock Springer US.

\bibitem[\protect\citeauthoryear{Blatman and Sudret}{Blatman and
  Sudret}{2011}]{BlatmanSudret}
Blatman, G. and B.~Sudret (2011).
\newblock Adaptive sparse polynomial chaos expansion based on least angle
  regression.
\newblock {\em Journal of computational Physics\/}~{\em 230}, 2345--2367.

\bibitem[\protect\citeauthoryear{Boyd, Parikh, Chu, Peleato, and Eckstein}{Boyd
  et~al.}{2011}]{10.1561/2200000016}
Boyd, S., N.~Parikh, E.~Chu, B.~Peleato, and J.~Eckstein (2011, January).
\newblock {Distributed Optimization and Statistical Learning via the
  Alternating Direction Method of Multipliers}.
\newblock {\em Found. Trends Mach. Learn.\/}~{\em 3\/}(1), 1--122.

\bibitem[\protect\citeauthoryear{Bubeck}{Bubeck}{2015}]{10.1561/2200000050}
Bubeck, S. (2015, November).
\newblock {Convex Optimization: Algorithms and Complexity}.
\newblock {\em Found. Trends Mach. Learn.\/}~{\em 8\/}(3-4), 231--357.

\bibitem[\protect\citeauthoryear{Carnell}{Carnell}{2021}]{lhspackage}
Carnell, R. (2021).
\newblock {\em lhs: Latin Hypercube Samples}.
\newblock R package version 1.1.3.

\bibitem[\protect\citeauthoryear{Dancik and Dorman}{Dancik and
  Dorman}{2008}]{mlegp}
Dancik, G.~M. and K.~S. Dorman (2008).
\newblock mlegp: statistical analysis for computer models of biological systems
  using {R}.
\newblock {\em Bioinformatics\/}~{\em 24\/}(17), 1966.

\bibitem[\protect\citeauthoryear{Durrande, Ginsbourger, and Roustant}{Durrande
  et~al.}{2012}]{Durrande2012}
Durrande, N., D.~Ginsbourger, and O.~Roustant (2012, 4).
\newblock {Additive Covariance kernels for high-dimensional Gaussian Process
  modeling}.
\newblock {\em Annales de la faculté des sciences de Toulouse
  Mathématiques\/}~{\em 21\/}(3), 481--499.

\bibitem[\protect\citeauthoryear{Durrande, Ginsbourger, Roustant, and
  Carraro}{Durrande et~al.}{2013}]{DURRANDE201357}
Durrande, N., D.~Ginsbourger, O.~Roustant, and L.~Carraro (2013).
\newblock {ANOVA} kernels and {RKHS} of zero mean functions for model-based
  sensitivity analysis.
\newblock {\em Journal of Multivariate Analysis\/}~{\em 115}, 57 -- 67.

\bibitem[\protect\citeauthoryear{Eddelbuettel}{Eddelbuettel}{2013}]{Eddelbuettel:2013:SRC:2517725}
Eddelbuettel, D. (2013).
\newblock {\em {Seamless R and C++ Integration with Rcpp}}.
\newblock Springer Publishing Company, Incorporated.

\bibitem[\protect\citeauthoryear{Eddelbuettel and Francois}{Eddelbuettel and
  Francois}{2019}]{RcppGSL}
Eddelbuettel, D. and R.~Francois (2019).
\newblock {\em RcppGSL: 'Rcpp' Integration for 'GNU GSL' Vectors and Matrices}.
\newblock R package version 0.3.7.

\bibitem[\protect\citeauthoryear{Fu}{Fu}{1998}]{10.2307/1390712}
Fu, W.~J. (1998).
\newblock Penalized {R}egressions: The {B}ridge versus the {L}asso.
\newblock {\em Journal of Computational and Graphical Statistics\/}~{\em
  7\/}(3), 397--416.

\bibitem[\protect\citeauthoryear{Galassi, Davies, Theiler, Gough, Jungman,
  Alken, Booth, Rossi, and Ulerich}{Galassi
  et~al.}{2018}]{galassi2018scientific}
Galassi, M., J.~Davies, J.~Theiler, B.~Gough, G.~Jungman, P.~Alken, M.~Booth,
  F.~Rossi, and R.~Ulerich (2018).
\newblock {GNU Scientific Library Reference Manual}.

\bibitem[\protect\citeauthoryear{Giraud}{Giraud}{2014}]{giraud2014introduction}
Giraud, C. (2014).
\newblock {\em Introduction to High-Dimensional Statistics}.
\newblock Chapman \& Hall/CRC Monographs on Statistics \& Applied Probability.
  Taylor \& Francis.

\bibitem[\protect\citeauthoryear{Gu}{Gu}{2019}]{10.1214/18-BA1133}
Gu, M. (2019).
\newblock {Jointly Robust Prior for Gaussian Stochastic Process in Emulation,
  Calibration and Variable Selection}.
\newblock {\em Bayesian Analysis\/}~{\em 14\/}(3), 857 -- 885.

\bibitem[\protect\citeauthoryear{Gu, Palomo, and Berger}{Gu
  et~al.}{2019}]{Gu_2019}
Gu, M., J.~Palomo, and O.~Berger, James (2019).
\newblock {RobustGaSP: Robust Gaussian Stochastic Process Emulation in R}.
\newblock {\em The R Journal\/}~{\em 11\/}(1), 112.

\bibitem[\protect\citeauthoryear{Gu, Wang, and Berger}{Gu
  et~al.}{2018}]{10.1214/17-AOS1648}
Gu, M., X.~Wang, and J.~O. Berger (2018).
\newblock {Robust Gaussian stochastic process emulation}.
\newblock {\em The Annals of Statistics\/}~{\em 46\/}(6A), 3038 -- 3066.

\bibitem[\protect\citeauthoryear{Guennebaud, Jacob, et~al.}{Guennebaud
  et~al.}{2010}]{eigenweb}
Guennebaud, G., B.~Jacob, et~al. (2010).
\newblock Eigen v3.
\newblock http://eigen.tuxfamily.org.

\bibitem[\protect\citeauthoryear{Hastie, Tibshirani, and Wainwright}{Hastie
  et~al.}{2015}]{Hastie:2015:SLS:2834535}
Hastie, T., R.~Tibshirani, and M.~Wainwright (2015).
\newblock {\em Statistical Learning with Sparsity: The Lasso and
  Generalizations}.
\newblock Chapman \& Hall/CRC.

\bibitem[\protect\citeauthoryear{Homma and Saltelli}{Homma and
  Saltelli}{1996}]{HOMMA19961}
Homma, T. and A.~Saltelli (1996).
\newblock Importance measures in global sensitivity analysis of nonlinear
  models.
\newblock {\em Reliability Engineering \& System Safety\/}~{\em 52\/}(1), 1 --
  17.

\bibitem[\protect\citeauthoryear{Huet and Taupin}{Huet and
  Taupin}{2017}]{huet:hal-01434895}
Huet, S. and M.-L. Taupin (2017).
\newblock Metamodel construction for sensitivity analysis.
\newblock {\em ESAIM: Procs\/}~{\em 60}, 27--69.

\bibitem[\protect\citeauthoryear{Kamari}{Kamari}{2019}]{rkhsmeamodpackage}
Kamari, H. (2019).
\newblock {\em RKHSMetaMod: Ridge Group Sparse Optimization Problem for
  Estimation of a Meta Model Based on Reproducing Kernel Hilbert Spaces}.
\newblock R package version 1.1.

\bibitem[\protect\citeauthoryear{Kennedy and O'Hagan}{Kennedy and
  O'Hagan}{2000}]{Kennedy00bayesiancalibration}
Kennedy, M.~C. and A.~O'Hagan (2000).
\newblock {Bayesian Calibration of Computer Models}.
\newblock {\em Journal of the Royal Statistical Society, Series B,
  Methodological\/}~{\em 63}, 425--464.

\bibitem[\protect\citeauthoryear{Kimeldorf and Wahba}{Kimeldorf and
  Wahba}{1970}]{kimeldorf1970}
Kimeldorf, G.~S. and G.~Wahba (1970, 04).
\newblock {A Correspondence Between Bayesian Estimation on Stochastic Processes
  and Smoothing by Splines}.
\newblock {\em Ann. Math. Statist.\/}~{\em 41\/}(2), 495--502.

\bibitem[\protect\citeauthoryear{Kleijnen}{Kleijnen}{2007}]{Kleijnen:2007:DAS:1554802}
Kleijnen, J. P.~C. (2007).
\newblock {\em Design and Analysis of Simulation Experiments\/} (1st ed.).
\newblock Springer Publishing Company, Incorporated.

\bibitem[\protect\citeauthoryear{Kleijnen}{Kleijnen}{2009}]{KLEIJNEN2009707}
Kleijnen, J. P.~C. (2009).
\newblock Kriging metamodeling in simulation: A review.
\newblock {\em European Journal of Operational Research\/}~{\em 192\/}(3), 707
  -- 716.

\bibitem[\protect\citeauthoryear{Le~Gratiet, Cannamela, and Iooss}{Le~Gratiet
  et~al.}{2014}]{doi:10.1137/130926869}
Le~Gratiet, L., C.~Cannamela, and B.~Iooss (2014).
\newblock {A Bayesian Approach for Global Sensitivity Analysis of
  (Multifidelity) Computer Codes}.
\newblock {\em SIAM/ASA Journal on Uncertainty Quantification\/}~{\em 2\/}(1),
  336--363.

\bibitem[\protect\citeauthoryear{Le~Gratiet, Marelli, and Sudret}{Le~Gratiet
  et~al.}{2017}]{LeGratiet2017}
Le~Gratiet, L., S.~Marelli, and B.~Sudret (2017).
\newblock {\em Metamodel-Based Sensitivity Analysis: Polynomial Chaos
  Expansions and Gaussian Processes}, pp.\  1289--1325.
\newblock Cham: Springer International Publishing.

\bibitem[\protect\citeauthoryear{Lin and Zhang}{Lin and Zhang}{2006}]{lin2006}
Lin, Y. and H.~H. Zhang (2006, 10).
\newblock Component selection and smoothing in multivariate nonparametric
  regression.
\newblock {\em Ann. Statist.\/}~{\em 34\/}(5), 2272--2297.

\bibitem[\protect\citeauthoryear{Liu, Wasserman, and Lafferty}{Liu
  et~al.}{2009}]{NIPS2008_3616}
Liu, H., L.~Wasserman, and J.~D. Lafferty (2009).
\newblock Nonparametric regression and classification with joint sparsity
  constraints.
\newblock In D.~Koller, D.~Schuurmans, Y.~Bengio, and L.~Bottou (Eds.), {\em
  Advances in Neural Information Processing Systems 21}, pp.\  969--976. Curran
  Associates, Inc.

\bibitem[\protect\citeauthoryear{Marrel, Iooss, Laurent, and Roustant}{Marrel
  et~al.}{2009}]{MARREL2009742}
Marrel, A., B.~Iooss, B.~Laurent, and O.~Roustant (2009).
\newblock {Calculations of Sobol indices for the Gaussian process metamodel}.
\newblock {\em Reliability Engineering \& System Safety\/}~{\em 94\/}(3), 742
  -- 751.

\bibitem[\protect\citeauthoryear{Mebane and Sekhon}{Mebane and
  Sekhon}{2011}]{JSSv042i11}
Mebane, W. and J.~Sekhon (2011).
\newblock {Genetic Optimization Using Derivatives: The rgenoud Package for R}.
\newblock {\em Journal of Statistical Software, Articles\/}~{\em 42\/}(11),
  1--26.

\bibitem[\protect\citeauthoryear{Meier}{Meier}{2020}]{grppkg}
Meier, L. (2020).
\newblock {\em grplasso: Fitting User-Specified Models with Group Lasso
  Penalty}.
\newblock R package version 0.4-7.

\bibitem[\protect\citeauthoryear{Meier, van~de Geer, and B\"uhlmann}{Meier
  et~al.}{2008}]{MeiGeeBue08}
Meier, L., S.~van~de Geer, and P.~B\"uhlmann (2008).
\newblock The group lasso for logistic regression.
\newblock {\em Journal of the Royal Statistical Society. Series B\/}~{\em
  70\/}(1), 53--71.

\bibitem[\protect\citeauthoryear{Meier, van~de Geer, and B\"uhlmann}{Meier
  et~al.}{2009}]{meier2009}
Meier, L., S.~van~de Geer, and P.~B\"uhlmann (2009, 12).
\newblock High-dimensional additive modeling.
\newblock {\em Ann. Statist.\/}~{\em 37\/}(6B), 3779--3821.

\bibitem[\protect\citeauthoryear{Oakley and O'Hagan}{Oakley and
  O'Hagan}{2004}]{doi:10.1111/j.1467-9868.2004.05304.x}
Oakley, J.~E. and A.~O'Hagan (2004).
\newblock {Probabilistic sensitivity analysis of complex models: a Bayesian
  approach}.
\newblock {\em Journal of the Royal Statistical Society: Series B (Statistical
  Methodology)\/}~{\em 66\/}(3), 751--769.

\bibitem[\protect\citeauthoryear{Raskutti, Wainwright, and Yu}{Raskutti
  et~al.}{2009}]{NIPS2009_3688}
Raskutti, G., M.~J. Wainwright, and B.~Yu (2009).
\newblock Lower bounds on minimax rates for nonparametric regression with
  additive sparsity and smoothness.
\newblock In {\em Advances in Neural Information Processing Systems}.

\bibitem[\protect\citeauthoryear{Raskutti, Wainwright, and Yu}{Raskutti
  et~al.}{2012}]{Raskutti:2012:MRS:2503308.2188398}
Raskutti, G., M.~J. Wainwright, and B.~Yu (2012, February).
\newblock {Minimax-optimal Rates for Sparse Additive Models over Kernel Classes
  via Convex Programming}.
\newblock {\em J. Mach. Learn. Res.\/}~{\em 13\/}(1), 389--427.

\bibitem[\protect\citeauthoryear{Ravikumar, Lafferty, Liu, and
  Wasserman}{Ravikumar et~al.}{2009}]{Ravikumar}
Ravikumar, P., J.~Lafferty, H.~Liu, and L.~Wasserman (2009).
\newblock Sparse additive models.
\newblock {\em Journal of the Royal Statistical Society: Series B (Statistical
  Methodology)\/}~{\em 71\/}(5), 1009--1030.

\bibitem[\protect\citeauthoryear{Roustant, Ginsbourger, and Deville}{Roustant
  et~al.}{2012}]{JSSv051i01}
Roustant, O., D.~Ginsbourger, and Y.~Deville (2012).
\newblock {DiceKriging, DiceOptim: Two R Packages for the Analysis of Computer
  Experiments by Kriging-Based Metamodeling and Optimization}.
\newblock {\em Journal of Statistical Software, Articles\/}~{\em 51\/}(1),
  1--55.

\bibitem[\protect\citeauthoryear{Saitoh}{Saitoh}{1988}]{saitoh1988theory}
Saitoh, S. (1988).
\newblock {\em Theory of reproducing kernels and its applications}.
\newblock Pitman research notes in mathematics series. Longman Scientific \&
  Technical.

\bibitem[\protect\citeauthoryear{Saltelli}{Saltelli}{2002}]{SALTELLI2002280}
Saltelli, A. (2002).
\newblock Making best use of model evaluations to compute sensitivity indices.
\newblock {\em Computer Physics Communications\/}~{\em 145\/}(2), 280 -- 297.

\bibitem[\protect\citeauthoryear{Saltelli, Chan, and Scott}{Saltelli
  et~al.}{2009}]{saltelli2009sensitivity}
Saltelli, A., K.~Chan, and E.~Scott (2009).
\newblock {\em Sensitivity Analysis}.
\newblock Wiley.

\bibitem[\protect\citeauthoryear{Saltelli, Tarantola, and Chan}{Saltelli
  et~al.}{1999}]{doi:10.1080/00401706.1999.10485594}
Saltelli, A., S.~Tarantola, and K.~P.-S. Chan (1999).
\newblock {A Quantitative Model-Independent Method for Global Sensitivity
  Analysis of Model Output}.
\newblock {\em Technometrics\/}~{\em 41\/}(1), 39--56.

\bibitem[\protect\citeauthoryear{Schoutens}{Schoutens}{2000}]{schoutens2000stochastic}
Schoutens, W. (2000).
\newblock {\em Stochastic Processes and Orthogonal Polynomials}.
\newblock Lecture Notes in Statistics. Springer New York.

\bibitem[\protect\citeauthoryear{Sobol}{Sobol}{1993}]{Sobol1993SensitivityEF}
Sobol, I.~M. (1993).
\newblock {Sensitivity Estimates for Nonlinear Mathematical Models}.
\newblock In {\em Sensitivity Estimates for Nonlinear Mathematical Models}.

\bibitem[\protect\citeauthoryear{Soize and Ghanem}{Soize and
  Ghanem}{2004}]{doi:10.1137/S1064827503424505}
Soize, C. and R.~Ghanem (2004).
\newblock {Physical Systems with Random Uncertainties: Chaos Representations
  with Arbitrary Probability Measure}.
\newblock {\em SIAM Journal on Scientific Computing\/}~{\em 26\/}(2), 395--410.

\bibitem[\protect\citeauthoryear{Sudret}{Sudret}{2008}]{SUDRET2008964}
Sudret, B. (2008).
\newblock Global sensitivity analysis using polynomial chaos expansions.
\newblock {\em Reliability Engineering \& System Safety\/}~{\em 93\/}(7), 964
  -- 979.
\newblock Bayesian Networks in Dependability.

\bibitem[\protect\citeauthoryear{Van~der Vaart}{Van~der
  Vaart}{1998}]{vaart_1998}
Van~der Vaart, A.~W. (1998).
\newblock {\em Asymptotic Statistics}.
\newblock Cambridge Series in Statistical and Probabilistic Mathematics.
  Cambridge University Press.

\bibitem[\protect\citeauthoryear{Welch, Buck, Sacks, Wynn, Mitchell, and
  Morris}{Welch et~al.}{1992}]{10.2307/1269548}
Welch, W.~J., R.~J. Buck, J.~Sacks, H.~P. Wynn, T.~J. Mitchell, and M.~D.
  Morris (1992).
\newblock {Screening, Predicting, and Computer Experiments}.
\newblock {\em Technometrics\/}~{\em 34\/}(1), 15--25.

\bibitem[\protect\citeauthoryear{Wiener}{Wiener}{1938}]{10.2307/2371268}
Wiener, N. (1938).
\newblock {The Homogeneous Chaos}.
\newblock {\em American Journal of Mathematics\/}~{\em 60\/}(4), 897--936.

\bibitem[\protect\citeauthoryear{Yang and Zou}{Yang and
  Zou}{2015}]{Yang:2015:FUA:2833490.2833520}
Yang, Y. and H.~Zou (2015, November).
\newblock {A Fast Unified Algorithm for Solving Group-lasso Penalize Learning
  Problems}.
\newblock {\em Statistics and Computing\/}~{\em 25\/}(6), 1129--1141.

\bibitem[\protect\citeauthoryear{Yang, Zou, and Bhatnagar}{Yang
  et~al.}{2020}]{gglspkg}
Yang, Y., H.~Zou, and S.~Bhatnagar (2020).
\newblock {\em gglasso: Group Lasso Penalized Learning Using a Unified BMD
  Algorithm}.
\newblock R package version 1.5.

\bibitem[\protect\citeauthoryear{Yuan and Lin}{Yuan and
  Lin}{2006}]{doi:10.1111/j.1467-9868.2005.00532.x}
Yuan, M. and Y.~Lin (2006).
\newblock Model selection and estimation in regression with grouped variables.
\newblock {\em Journal of the Royal Statistical Society: Series B (Statistical
  Methodology)\/}~{\em 68\/}(1), 49--67.

\end{thebibliography}
\end{document}